\documentclass[pdflatex,sn-vancouver-num]{sn-jnl}

\usepackage{graphicx}%
\usepackage{multirow}%
\usepackage{amsmath,amssymb,amsfonts}%
\usepackage{amsthm}%
\usepackage{mathrsfs}%
\usepackage[title]{appendix}%
\usepackage{xcolor}%
\usepackage{colortbl}%
\usepackage{textcomp}%
\usepackage{manyfoot}%
\usepackage{booktabs}%
\usepackage{algorithm}%
\usepackage{algorithmicx}%
\usepackage{algpseudocode}%
\usepackage{listings}%
\usepackage{mathtools}%
\usepackage{soul}%
\usepackage{framed}%

\providecommand{\FP}{\mathrm{FP}}

\definecolor{revhl}{RGB}{255,225,70}
\definecolor{taskrow}{gray}{0.9}
\sethlcolor{revhl}
\colorlet{shadecolor}{revhl}
\soulregister{\ref}{1}
\soulregister{\autoref}{1}
\soulregister{\cite}{1}
\soulregister{\citep}{1}
\soulregister{\citet}{1}
\sethlcolor{revhl}

\makeatletter
 {\par\unskip\endMakeFramed}
\makeatother

\theoremstyle{thmstyleone}%
\theoremstyle{thmstyletwo}%

\theoremstyle{thmstylethree}%

\begin{document}

\title[PGFS++: Molecular Property Improvement under Synthesis and Diversity Constraints]{PGFS++: Molecular Property Improvement\\under Synthesis and Diversity Constraints}

\author[1]{\fnm{Boqiao} \sur{Zhang}}

\author[1]{\fnm{Godbless Tamaraebi} \sur{James}}

\author{\fnm{Sai Krishna} \sur{Gottipati}}

\author*[1,2]{\fnm{Andrew} \sur{Fitzgibbon}}\email{awf@graphcore.ai}

\affil*[1]{\orgname{University of Cambridge}, \orgaddress{\country{United Kingdom}}}

\affil[2]{\orgname{Graphcore}, \orgaddress{\country{United Kingdom}}}

\abstract{Improving molecular properties, such as drug-likeness or binding affinity, is a recurring task in early-stage drug discovery. However, molecules optimized in an unconstrained chemical space have limited practical value if they cannot be synthesized.
Policy Gradient for Forward Synthesis (PGFS) is a synthesis-aware reinforcement learning (RL) method for molecular improvement, but its use of reactant embedding prediction makes reactant selection indirect, which, as we show, limits learning effectiveness.
We first develop PGFS+, in which reaction templates and second reactants are represented by trainable embedding lookup tables. Combined with a more effective scoring function and RL algorithm, PGFS+ significantly improves the desired property. However, it exposes a reward-hacking failure mode: a powerful reactant search can map diverse input molecules to the same high-reward \lq\lq magnet molecule\rq\rq, improving the reward while collapsing the output diversity.
We therefore introduce PGFS++, a synthesis-aware reinforcement learning framework for input-specific molecular improvement. Given an input molecule, PGFS++ treats it as the start of a forward-synthesis trajectory, applies learned reaction templates with compatible in-stock building blocks, and produces a molecule with improved target properties, an explicit synthesis route, and structural similarity to the input.
Experiments on molecular improvement tasks show that PGFS++ improves target properties while preserving high output diversity.

\textbf{Scientific Contribution}
We found that the indirect reactant selection mechanism in PGFS limits learning
effectiveness, despite avoiding the need to score a large discrete action space.
PGFS+ directly scores discrete reaction-template and second-reactant choices using an attention-like mechanism, with a precomputed compatibility mask to reduce computational overhead. We further identify magnet-molecule convergence as a reward-hacking failure mode in PGFS+ and address it in PGFS++ by coupling synthesis-constrained property optimisation with input-specific molecular improvement.
}

\keywords{Molecular Optimization; Drug Discovery; Reinforcement Learning; Machine Learning; Forward Synthesis; Synthesis Constraints}

\maketitle

\section{Introduction}
\label{sec:introduction}

Designing novel molecules has the potential to drive advances across fields such as medicine and materials science. Conventional drug discovery workflows are generally driven via iterative design–make–purify–test cycles, with decisions guided largely by chemical expertise \citep{chow2018streamlining}. Although valuable, traditional methods are often resource-intensive and can only explore a very limited fraction of the chemical space, which encompasses all possible organic molecules and is estimated to contain approximately $10^{23}$ to $10^{60}$ potential compounds \citep{COLEY2021133}. 
To explore this vast chemical space more efficiently, artificial intelligence (AI)-based approaches to chemistry and drug discovery have advanced substantially over recent decades, offering a promising way to accelerate early-stage drug discovery. 

A central distinction among AI methods for molecular optimisation is whether synthesis constrains candidate generation. Methods without explicit synthesis constraints can optimise molecular properties without specifying how the resulting candidates could be synthesised. These methods include genetic algorithms\citep{revintro_brown2004}, variational autoencoders \citep{revintro_gomez2018}, sequence-based generative systems \citep{reinvent4}, and diffusion models \citep{revintro_vignac2023}. While these methods can generate potentially helpful molecules, they usually requires additional retrosynthesis model or expert assessment to determine whether the generated structure can be practically synthesized. Synthesis-constrained methods address this limitation by constructing products from available reaction templates and building blocks. Within this family, Policy Gradient for Forward Synthesis (PGFS) uses reinforcement learning (RL) to select reactions and reactants \citep{pgfs}, while Reaction-GFlowNet (RGFN) and SynFlowNet use generative flow networks to explore reaction-defined chemical spaces \citep{revintro_koziarski2024,revintro_cretu2025}. SynNet combines evolutionary search with learned synthesis-tree generation, and SyntheMol uses Monte Carlo tree search \citep{revintro_gao2022,synthemol}. ChemProjector and SynFormer use transformer-based models of synthesis pathways, while SynLlama uses a large language model to generate such pathways \citep{revintro_luo2024,revintro_gao2025,revintro_sun2025}. SynCoGen further incorporates three-dimensional molecular coordinates into reaction-constrained generation \citep{revintro_rekesh2026}. These approaches integrate reaction constraints into molecular generation, yielding candidates optimised for desired properties together with proposed synthetic routes.

Beyond target properties, evaluating molecular optimisation involves assessing synthetic accessibility, similarity to specified starting molecules and diversity among generated products. Synthetic complexity or accessibility can be estimated using measures such as the synthetic accessibility (SA) score (SAscore), the synthetic complexity score (SCScore), the retrosynthetic accessibility score (RAscore) and DeepSA, but these measures cannot independently validate a proposed synthesis route \citep{revintro_ertl2009,revintro_coley2018,revintro_thakkar2021,revintro_wang2023}. For input-conditioned tasks, including reaction-aware optimisation studied in Syn-MolOpt and TRACER, input–output similarity describes how closely each product resembles its starting molecule \citep{revintro_yin2025,revintro_nakamura2025}. Output diversity, by contrast, describes how much the generated products differ from one another. A property-driven model can map many different inputs to the same high-scoring product, which reduces both input specificity and output diversity, even when the generated routes comply with the allowed reaction templates.

PGFS is a widely used reinforcement learning framework for synthesis-constrained molecular optimisation. It selects the second reactant by predicting a continuous representation and retrieving a building block through nearest-neighbour search \citep{pgfs}. However, its performance has become less competitive as more advanced methods have emerged. We find that its indirect reactant selection mechanism limits learning effectiveness, despite avoiding the need to score a large discrete action space. In this work, our main contributions are:

\begin{enumerate}
    \item We present PGFS+, which replaces continuous reactant prediction and nearest-neighbour retrieval with a categorical policy over template-compatible second reactants. Compatibility masking makes this discrete action space computationally tractable, and our experiments show that PGFS+ achieves higher property rewards than PGFS.

    \item We identify the \emph{magnet-output problem}: PGFS+ can map many distinct starting molecules to the same high-scoring product, reducing both output diversity and input-specificity. Our experiments show that the diversity-encouraging strategies we evaluated do not adequately prevent this collapse. This finding identifies a failure mode of property-driven molecular optimisation in the PGFS framework.

    \item We present PGFS++, which retains direct reactant selection while adding an input--output similarity bonus and a reward cap below a similarity threshold. We derive an upper bound on the reduction in output diversity and show how increasing input--output similarity tightens this bound. Our experiments show that PGFS++ mitigates magnet-output collapse while retaining most of the property gains obtained by PGFS+.
\end{enumerate}

\section{Methods}
\label{sec:methods}

\subsection{Overview}
Following PGFS, we use an input building block as the first reactant in each episode, and represent molecular improvement as a sequence of unimolecular or bimolecular reactions applied to the input building block. Each synthetic step is decomposed into two sub-actions: selecting a reaction template and, for bi-molecular reactions, selecting a compatible second reactant from the set of available building blocks. The reaction template is chosen from templates available in laboratory or industrial synthesis, and the second reactant is selected from a set of available building blocks.

The reinforcement learning agent first learns to select an appropriate template $T$ given the current reactant $R_1$. Conditioned on $R_1$ and $T$, it then selects a second reactant $R_2$ to maximize the reward associated with the properties of the product molecule. This hierarchical action decomposition ensures that the generated molecules are associated with explicit synthetic routes and improves training efficiency by reducing the complexity of each decision.

A key challenge is that there may be millions of possible $R_2$ candidates for each $T$ and $R_1$. PGFS addresses this by using a neural network to predict an $R_2$ embedding, followed by $k$-nearest-neighbor retrieval from the building-block set. 

In contrast, PGFS+ and PGFS++ directly scores feasible $R_2$ candidates after applying a template-specific compatibility mask. We observed that the large $R_2$ action space can be efficiently reduced to a tractable candidate set using RDKit (version 2026.3.3) substructure matching. We pre-compute a $T$-$R_2$ compatibility mask for each bimolecular template. For our dataset, which contains approximately 118,000 molecules and 87 bimolecular templates, pre-computing the masks for all bimolecular templates takes approximately 10 minutes on a single CPU process. This computation is performed only once and can be reused for subsequent training runs. After masking, each bimolecular template has a median of approximately 1000 compatible $R_2$ candidates and a maximum of 55,000, making direct attention-like scoring feasible. Although the compatibility mask cannot guarantee successful reaction execution or product validity in RDKit, it excludes second reactants that fail the template's substructure requirements. We assessed the reliability of mask-permitted candidates as below. For each of the {$87$} bimolecular templates, we sampled compatible first reactants and mask-permitted second reactants and attempted up to {$1{,}000$} reactions and all reactions formed from these combinations returned a valid product. Training records provided consistent evidence in Table~\ref{tab:reaction-counts} where each environment step was classified as an invalid reaction, a STOP action or a successful reaction. As the table shows, no invalid reactions were recorded for models with different random seeds on the quantitative estimate of drug-likeness (QED), soluble epoxide hydrolase (sEH) and human immunodeficiency virus (HIV) integrase (HIV-Int) tasks (Table~\ref{tab:reaction-counts}). If a mask-permitted reaction fails, the step receives the invalid-reaction penalty, set to {$-1$} by default in our implementation, and the transition is retained in the proximal policy optimization (PPO) update.

\begin{table}[htbp]
\caption{Environment steps during PGFS++ training with pre-computed $T$-$R_2$ compatible masks. Invalid reactions are actions for which RDKit returned no valid product. Models 1, 2, and 3 are the three independently trained runs of each objective. }\label{tab:reaction-counts}
\centering
\small
\setlength{\tabcolsep}{7pt}
\renewcommand{\arraystretch}{1.15}
\begin{tabular}{ccccc}
\toprule
Objective & Model & Invalid reaction & STOP & Successful reactions \\
\midrule
\multirow{3}{*}{QED}
 & 1 & 0 & 532{,}277 & 469{,}195 \\
 & 2 & 0 & 436{,}716 & 564{,}756 \\
 & 3 & 0 & 531{,}135 & 470{,}337 \\
\addlinespace
\multirow{3}{*}{sEH}
 & 1 & 0 & 251{,}203 & 750{,}269 \\
 & 2 & 0 & 252{,}940 & 748{,}532 \\
 & 3 & 0 & 267{,}502 & 877{,}769 \\
\addlinespace
\multirow{3}{*}{HIV-integrase}
 & 1 & 0 & 245{,}835 & 755{,}637 \\
 & 2 & 0 & 264{,}257 & 737{,}215 \\
 & 3 & 0 & 304{,}659 & 696{,}813 \\
\bottomrule
\end{tabular}
\end{table}

\subsection{Masked hierarchical policy}
\label{sec:masked_policy}

\begin{figure}[htbp]
\centering
\includegraphics[width=0.8\textwidth,]{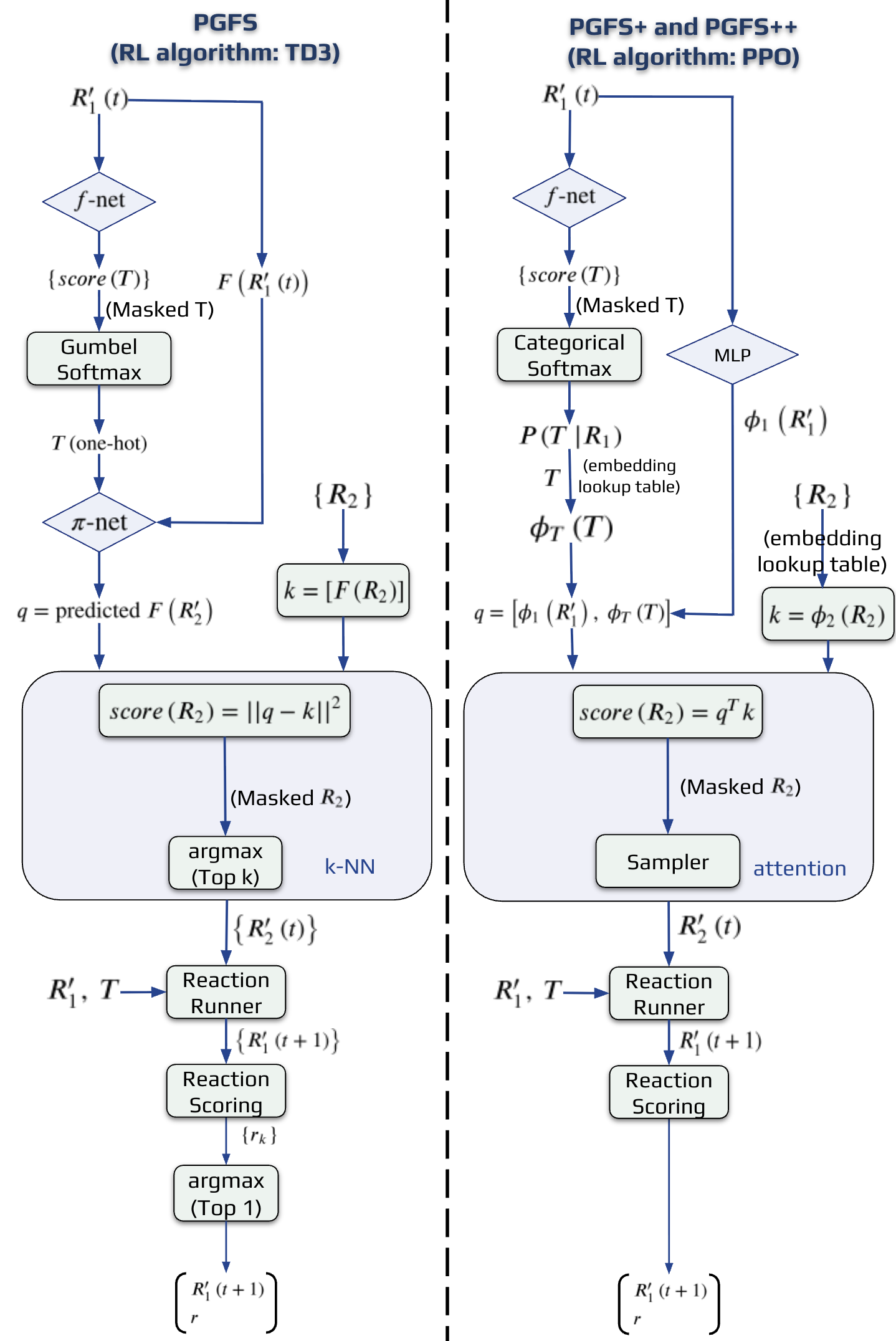}
\caption{Workflow comparison between PGFS, PGFS+, and PGFS++. At the first reaction step, the input molecule serves as the first reactant $R_1'(t)$. At each subsequent step, the product of the preceding reaction serves as $R_1'(t)$. At step $t$, both methods use an $f$-network conditioned on the current first reactant $R_1'(t)$ to predict the reaction template $T$. PGFS uses the fingerprint representation $F(R_1'(t))$ directly, whereas PGFS+ and PGFS++ first transform the $R_1$ representation through an additional multilayer perceptron (MLP). For second-reactant selection, PGFS represents $R_2$ candidates using fixed fingerprints and retrieves candidates by k-nearest neighbours (kNN). In contrast, PGFS+ and PGFS++ represent template-compatible $R_2$ candidates with a trainable embedding lookup table and score them globally using an attention-like mechanism with query $q=\mathrm{concat}(\phi_1(R_1'),\phi_T(T))$. Once $R_2'(t)$ is selected, the reaction runner applies the chosen reaction and returns the product, which becomes the next state $R_1'(t+1)$.}
\label{fig:workflow}
\end{figure}

PGFS+ and PGFS++ share the hierarchical policy architecture described below, with PGFS++ additionally incorporating input--output similarity shaping into the reward (Section~\ref{sec:method_in_out_sim}). In this architecture, the actor encodes the Morgan fingerprint of the current molecule $R_1$ into a learned representation $\phi_1(R_1)$ (Fig.~\ref{fig:workflow}). This representation conditions a two-stage action-selection process: selecting a reaction template or STOP, followed, for bimolecular templates, by selecting a compatible second reactant. To train this hierarchical categorical policy, we replace twin delayed deep deterministic policy gradient (TD3), the continuous-action reinforcement learning algorithm used in PGFS, with PPO, which naturally supports discrete action spaces.

\paragraph{Template masking and sampling.}
Template selection is restricted by a mask that identifies admissible reactions for the current molecule $R_1$. This mask retains templates whose first-reactant substructure patterns match $R_1$. For bimolecular templates, retention additionally requires at least one compatible second reactant in the precomputed building-block mask. The retained templates, together with the always-available STOP action, form the admissible set $\mathcal{T}(R_1)$. Over this set, the template logits $z_T$ are normalised to define the categorical policy
\[
\pi_T(T\mid R_1)
=
\frac{\exp(z_T)}
{\sum_{U\in\mathcal{T}(R_1)}\exp(z_U)},
\qquad T\in\mathcal{T}(R_1).
\]
This distribution assigns zero probability to excluded templates and is used to sample $T$. The sampled choice determines whether a second selection is needed: STOP and unimolecular templates complete action selection, whereas bimolecular templates require a second reactant.

\paragraph{Second-reactant masking and sampling.}
For a sampled bimolecular template $T$, the precomputed compatibility mask identifies the available second-reactant candidates. These candidates form an index set $I_T$ of size $K_T=|I_T|$, with each index referring to a building block and its learned key embedding. To score these key embeddings, the actor constructs a query from the representations of the current molecule and selected template:
\[
q=\mathrm{concat}\bigl(\phi_1(R_1),\phi_T(T)\bigr).
\]
The query scores only the keys indexed by $I_T$, producing a logit $s_i$ for each compatible candidate. These logits are normalised over $I_T$ to define the conditional second-reactant policy
\[
\pi_{R_2}(i\mid R_1,T)
=
\frac{\exp(s_i)}
{\sum_{j\in I_T}\exp(s_j)},
\qquad i\in I_T.
\]
From this distribution, the actor samples an index $i$ and selects the corresponding building block as $R_2$. The selection therefore involves exactly $K_T$ possible outcomes, with zero probability assigned to candidates outside $I_T$. Because $I_T$ is template-specific, the number of outcomes can vary across selected templates.

\paragraph{Action log-probabilities and PPO updates.}
The sampled template and second-reactant index jointly define a bimolecular action $a=(T,i)$. Its log-probability combines the template log-probability with the conditional log-probability of the sampled building-block index. This conditional term is computed by applying log-softmax over the same candidate set used for sampling:
\[
\log\pi_{R_2}(i\mid R_1,T)
=
s_i-\log\sum_{j\in I_T}\exp(s_j).
\]
Combining this term with the template log-probability gives the complete action log-probability. For STOP and unimolecular reactions, the action is simply $a=T$, and only the template term is required:
\[
\ell(a,R_1)
=
\begin{cases}
\log\pi_T(T\mid R_1),
\text{for STOP or a unimolecular template},\\[2pt]
\log\pi_T(T\mid R_1)
+\log\pi_{R_2}(i\mid R_1,T),
\text{for a bimolecular template}.
\end{cases}
\]

These action log-probabilities are evaluated under both the rollout policy and the updated policy to construct the PPO probability ratio. To ensure that both evaluations use the same admissible actions, each transition records its template mask and, where applicable, its second-reactant index set. During optimisation, these recorded masks and indices are used to recompute the sampled action's log-probability under the updated policy. The updated log-probability $\ell_{\theta}$ is compared with the fixed rollout log-probability $\ell_{\mathrm{old}}$ through
\[
\rho_t(\theta)
=
\exp\!\left[
\ell_{\theta}(a_t,R_{1,t})
-
\ell_{\mathrm{old}}(a_t,R_{1,t})
\right].
\]
This ratio enters the clipped PPO objective. Its computation accommodates different candidate-set sizes across transitions because both policies evaluate each sampled action over the same admissible set.

\subsection{Magnet-output collapse and input-output similarity}
\label{sec:method_in_out_sim}
As shown in Fig.~\ref{fig:magnet_molecule}, when PGFS+ is trained only with a target-property reward, such as $\Delta$QED, the policy can learn to map diverse input molecules to the same high-reward output molecule. Specifically, among 12,689 randomly selected test episodes, 12,415 (97.8\%) produced the same output molecule. Empirically, we found that the collapse occurs often because the policy tends to select similar templates 
and second reactants 
across different
starting molecules, producing similar or even identical high-reward outputs.
In other words, the model improves different inputs by pushing them toward a small number of high-scoring magnet molecules, rather than producing input-specific improved molecules. Although we identify this behaviour in PGFS+, it reflects a broader failure mode in synthesis-aware molecular improvement tasks: when the reward depends mainly on terminal molecular properties and does not constrain input--output relatedness, the policy can be incentivised to collapse diverse inputs toward a small set of high-reward products. To mitigate this problem, PGFS++ introduces input--output similarity shaping into the reward function, encouraging the output molecule of each episode to remain structurally close to its corresponding input molecule.

\begin{figure*}[!tp]
\centering
\includegraphics[width=\textwidth]{Figures/common_molecule_plot.pdf}
\caption{Reward hacking phenomenon in PGFS+. \textbf{A}, PGFS+ maps hundreds of input molecules to the same output molecule, causing high QED but low output diversity. This is referred to as the magnet output phenomenon. Highlighted atoms indicate the portions of the output molecule that are traced back to the corresponding input molecules. \textbf{B}, Trade-off between mean QED and mean diversity for PGFS, its variants, and PGFS+. Because of the magnet output, PGFS+ achieves high mean QED but has output diversity close to zero.}
\label{fig:magnet_molecule}
\end{figure*}

To discourage magnet-output collapse, PGFS++ makes similarity to the starting molecule an explicit part of the episode reward. We design the reward function incorporating input--output similarity bonus:
\begin{equation}
\label{eq:shaped}
\begin{aligned}r_{\mathrm{episode}}^{\times} = r_{\mathrm{property}}(m_{\mathrm{out}})
    \big(1 + c\,\mathrm{sim}(m_{\mathrm{in}},m_{\mathrm{out}})\big)
\end{aligned}
\end{equation}
where $r_{\mathrm{property}}$ is the property-based reward, such as 
$\Delta$QED and $\Delta$sEH, $c>0$ control the bonus strength, and
$\mathrm{sim}(m_{\mathrm{in}}, m_{\mathrm{out}})$ denotes the Tanimoto similarity \citep{tanimoto_sim} between input and output molecules per episode.

When the input--output similarity falls below a tolerance threshold $\tau$, we cap
the episode reward:
\begin{equation}
\label{eq:in_out_bonus_reward}
r_{\mathrm{episode}}
=
\begin{cases}
r_{\mathrm{episode}}^{\times},
& \mathrm{sim}(m_{\mathrm{in}},m_{\mathrm{out}})\ge \tau,\\
\min\{r_{\mathrm{episode}}^{\times},\kappa\},
& \mathrm{sim}(m_{\mathrm{in}},m_{\mathrm{out}})<\tau,
\end{cases}
\end{equation}
where $\kappa$ is the reward cap. 

The similarity-shaped reward can mitigate the collapse problem and yield high output diversity (see Supplementary Section S2 for details).

\section{Experiments}
\label{sec:experiments}


We evaluated the proposed framework using the same building-block set as PGFS \citep{pgfs}, comprising approximately 118,000 molecules.
These building blocks were combined with 102 reaction templates
derived from \citet{dingos_dataset}, including 15 unimolecular
and 87 bimolecular templates.
From the building-block collection, we randomly selected
2,000 $R_1$ reactants as test input building blocks where 
each input served as the starting molecule for an
optimisation episode. For the HIV-Int optimisation task, we additionally evaluated PGFS++ on a out-of-distribution test set of 200 input molecules randomly selected from ChEMBL.

This input-specific optimisation task was consistently used to compare PGFS++
with PGFS, the towered actor--critic variant PGFS-TAC~\citep{gottipati2021towered}, the max-Bellman variant PGFS-MB~\citep{gottipati2023maximumrewardformulationreinforcement}, REINVENT4 mol2mol, SynFlowNet
and SynFormer.
REINVENT4 mol2mol shares the input-conditioned formulation
but generates molecules without enforcing synthesis constraints
\citep{reinvent4}.
Synthesis constraints are incorporated in SynFlowNet, which
uses generative flow networks for de novo molecular generation.
SynFormer also targets synthesis-constrained de novo generation,
using transformer-based models.
These two de novo methods were evaluated here on improving
specified starting molecules, and their results should be interpreted in light of task adaptation. 

For SynFlowNet, we used the official training implementation,
changing only the target scorer, reaction-template set and
building-block collection to match our evaluation setting.
For SynFormer, we followed the official implementation but
used a batch size of 32 because of computational resource constraints. For REINVENT4, we approximately matched the oracle budget used for PGFS++ training.
This budget was implemented by fine-tuning the medium prior
model for 500 reinforcement-learning steps, with 2,048
oracle-scored molecules per step, giving approximately
1.02 million oracle calls.
The remaining baselines used their default training settings.


\subsection{Optimization objectives}
\label{sec:objectives}

The experiments considered three molecular optimization objectives: quantitative estimate of drug-likeness (QED), predicted binding to soluble epoxide hydrolase (sEH), and predicted inhibitory activity against human immunodeficiency virus (HIV) integrase (HIV-Int).

\noindent\textbf{Drug-likeness (QED).}
Quantitative estimate of drug-likeness (QED) \citep{qed} was computed using RDKit. The score ranges from 0 to 1, with higher values indicating greater drug-likeness.

\noindent\textbf{Soluble epoxide hydrolase (sEH).}
Binding to sEH was estimated using the pretrained message-passing predictor adopted by SynFlowNet. The network output was divided by 8, following the scaling used in that benchmark.

\noindent\textbf{Predicted HIV-integrase pIC$_{50}$.}
HIV-Int inhibitory potency was estimated using the prediction model introduced in PGFS \citep{pgfs}. Potency is expressed as $\mathrm{pIC}_{50}=-\log_{10}(\mathrm{IC}_{50})$, where $\mathrm{IC}_{50}$ is the half-maximal inhibitory concentration in molar units, so higher values indicate greater predicted potency. The model was trained on ChEMBL25 measurements for HIV integrase (CHEMBL3471) that report pIC$_{50}$. After structural standardization, desalting, aggregation of repeated measurements, and molecular-weight and outlier filtering, the curated dataset comprised 775 compounds. Each molecule was represented by 199 RDKit descriptors.  The predicted pIC$_{50}$ of a molecule was calculated as the mean of the 25 model predictions.
\noindent\textbf{AntiHIV-Pred.}
To complement the ensemble-based potency estimates, each molecule was also scored with AntiHIV-Pred, an external model that predicts HIV-1 integrase $\mathrm{pIC}_{50}$ from chemical structure~\citep{stolbov2020antihivpred}.

\subsection{Metrics}
\label{sec:metrics}
In addition to the optimization objectives, we evaluated generated molecules using diversity, synthetic accessibility (SA), input-output similarity, and ASKCOS (Automated System for Knowledge-based Continuous Organic Synthesis) plausibility.

\def\FP{\mathrm{FP}}
\textbf{Diversity} measures structural variation within the output set $O$. It is defined as one minus the average pairwise Tanimoto similarity between Morgan fingerprints:
\[
D(O) = 1 - \frac{1}{|P|} \sum_{m_i,m_j \in P} T(\FP(m_i), \FP(m_j)),
\]
where $P=\{(m_i,m_j): m_i,m_j \in O, i<j\}$ is the set of unordered molecule pairs, $T$ denotes Tanimoto similarity and $\FP$ denotes the Morgan fingerprint representation.

\textbf{Synthetic accessibility} was assessed using the Ertl--Schuffenhauer synthetic accessibility (SA) score implemented in RDKit \citep{revintro_ertl2009,rdkit}. Although the SA score does not fully capture route-level synthetic feasibility, we used it as a simple molecule-level metric to enable comparison with non-synthesis-aware methods. For visual consistency with the other metrics, we report $1-\mathrm{SA}$ in plots, so that higher values indicate better synthetic accessibility. While the SA score is widely used to estimate molecule-level synthetic accessibility, it does not fully capture the route-level information of synthesis-aware methods. ASKCOS route plausibility below addresses this limitation by incorporating information from the generation routes.

\textbf{Input-output similarity} measures the structural similarity between the input molecule $m^e_{\mathrm{in}}$ and the output molecule $m^e_{\mathrm{out}}$ for each episode $e$, averaged across episodes. 
For each episode $e$, we compute
\[
\mathrm{sim}_{\mathrm{in,out}}(e) = T(\FP(m^e_{\mathrm{in}}), \FP(m^e_{\mathrm{out}})).
\]
This metric is important for molecule improving tasks, where the optimized molecule should retain structural similarity to the starting molecule. 

\textbf{ASKCOS plausibility.}
We assessed the synthetic plausibility of generated routes using the ASKCOS fast-filter model~\citep{ASKCOS}. This assessment used each method's proposed reaction sequence directly, without reconstructing alternative pathways. It therefore applied only to methods that produced explicit synthesis routes; methods without reaction trajectories, such as REINVENT4 in the evaluated configuration, were reported as not applicable.

For each proposed route, ASKCOS assigns every reaction step a plausibility score based on its reactants and product. These step-level scores, denoted by $p_n\in[0,1]$, indicate greater predicted chemical plausibility at higher values. To summarise plausibility across the complete route, we aggregated the scores multiplicatively, following a similar approach in ~\citet{mrugalla2025route}. For an episode $e$ containing $N_e$ reaction steps, the route-level score is
\begin{equation}
\label{eq:askcos_route_plausibility}
\mathrm{Plausibility}(e)
=
\prod_{n=1}^{N_e} p_n.
\end{equation}
This aggregation incorporates every constituent reaction, with lower-scoring steps reducing the overall score. The overall score also depends on route length, because each additional factor can only maintain or decrease the product.

\subsection{Optimisation of predicted HIV-integrase inhibitory potency}

\begin{figure*}[!tp]
\centering
\includegraphics[width=\textwidth]{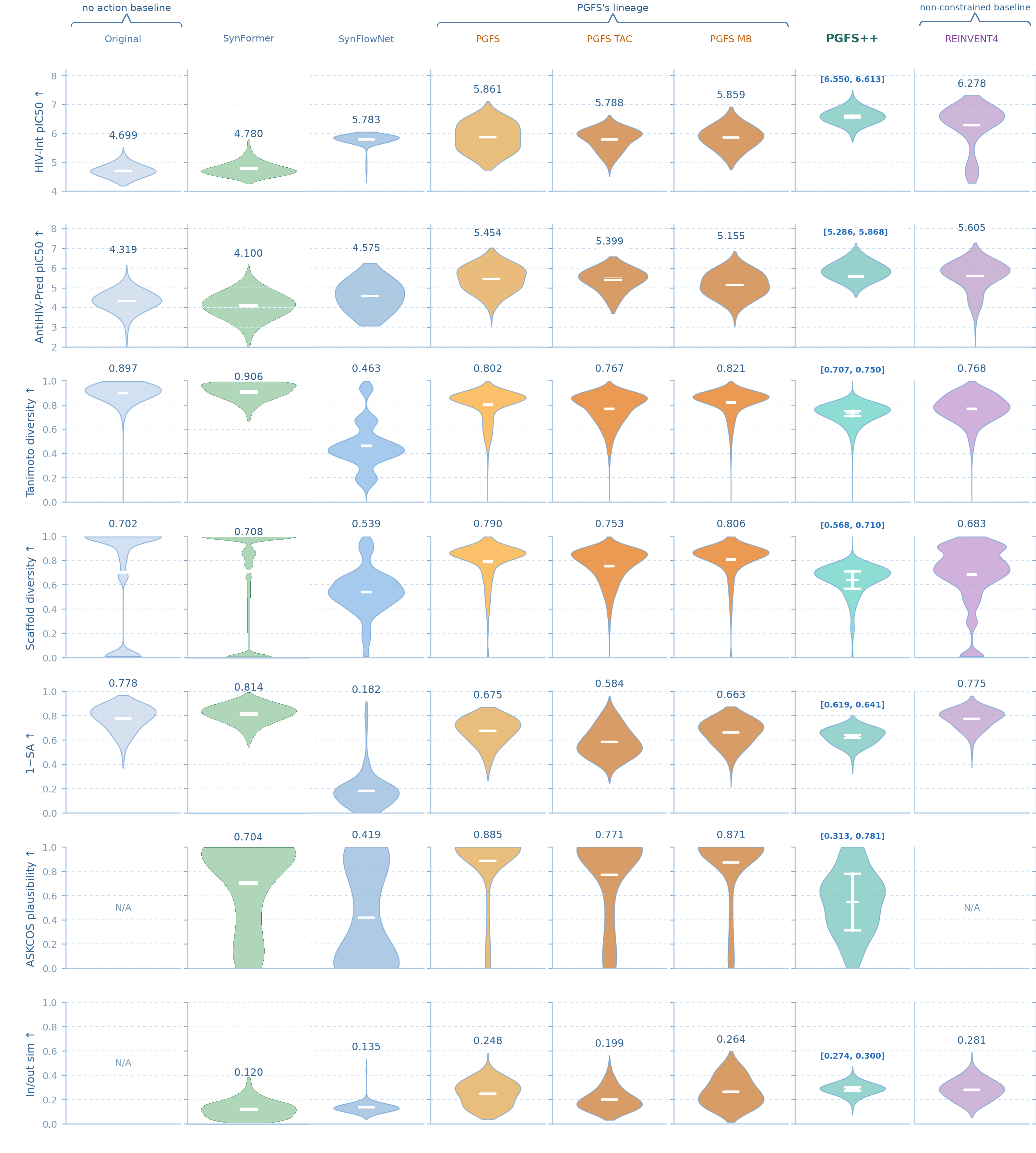}
\caption{Performance comparison on the HIV-Int improving task. AntiHIV-Pred pIC$_{50}$ is an external integrase model for independent evaluation. }
\label{fig:hiv_main_experiments}
\end{figure*}

HIV integrase (HIV-Int) catalyses the 3$^{\prime}$-processing and
strand-transfer reactions required to integrate viral DNA into
the host genome and is an established target for antiretroviral
treatment~\citep{pgfs}. For this target, we evaluated whether PGFS++ could improve
predicted inhibitory potency using the computational objective
defined in Section~\ref{sec:objectives}.

PGFS++ improved predicted HIV-Int potency under both predictors, with different advantages over REINVENT4 and SynFlowNet (Fig.~\ref{fig:hiv_main_experiments}). Under the HIV-Int ensemble, its seed-wise mean $\mathrm{pIC}_{50}$ ranged from 6.550 to 6.613, exceeding REINVENT4 (6.278), the PGFS baselines (5.788--5.861) and SynFlowNet (5.783). Although the PGFS++ and REINVENT4 distributions overlapped, REINVENT4 exhibited a longer tail towards low predicted potency. Under AntiHIV-Pred, PGFS++ achieved seed-wise means of 5.286--5.868, compared with 4.319 for the starting molecules. This range encompassed the REINVENT4 mean of 5.605, indicating comparable predicted potency with variation across PGFS++ seeds.
Beyond potency, PGFS++ and REINVENT4 achieved similar input--output similarity (0.274--0.300 versus 0.281), while differing in diversity and synthetic accessibility. PGFS++ had lower Tanimoto diversity (0.707--0.750 versus 0.768), whereas its scaffold-diversity range encompassed the REINVENT4 score (0.568--0.710 versus 0.683). Its $1-\mathrm{SA}$ scores were also lower (0.619--0.641 versus 0.775), indicating less favourable synthetic accessibility according to this metric. PGFS++, however, additionally provided explicit reaction routes, enabling assessment of their predicted plausibility.

Among the methods that generate explicit synthesis routes, PGFS++ achieved larger mean gains in predicted HIV-Int potency than SynFormer and SynFlowNet in the input optimisation setting. Under the HIV-Int ensemble, SynFormer reached a mean $\mathrm{pIC}_{50}$ of 4.780, an increase of 0.081 over the starting molecules. SynFlowNet reached a mean of 5.783, compared with PGFS++ seed-wise means of 6.550--6.613. Mean input--output Tanimoto similarity was 0.274--0.300 for PGFS++, compared with 0.120 for SynFormer and 0.135 for SynFlowNet. Relative to SynFlowNet, this greater retention of input-related features was accompanied by higher Tanimoto diversity (0.707--0.750 versus 0.463), scaffold diversity (0.568--0.710 versus 0.539), and $1-\mathrm{SA}$ scores (0.619--0.641 versus 0.182).
Still, these advantages in product-level metrics did not translate into consistently higher route plausibility. PGFS++ seed-wise mean ASKCOS scores ranged from 0.313 to 0.781, encompassing the scores of both SynFlowNet (0.419) and SynFormer (0.704). Thus, PGFS++ combined larger predicted potency gains with greater input--output similarity, while its predicted route plausibility varied substantially across seeds.

\subsubsection{Out-of-distribution test on ChEMBL molecules}
\label{sec:chembl-ood}

\begin{figure*}[!tp]
\centering
\includegraphics[width=\textwidth]{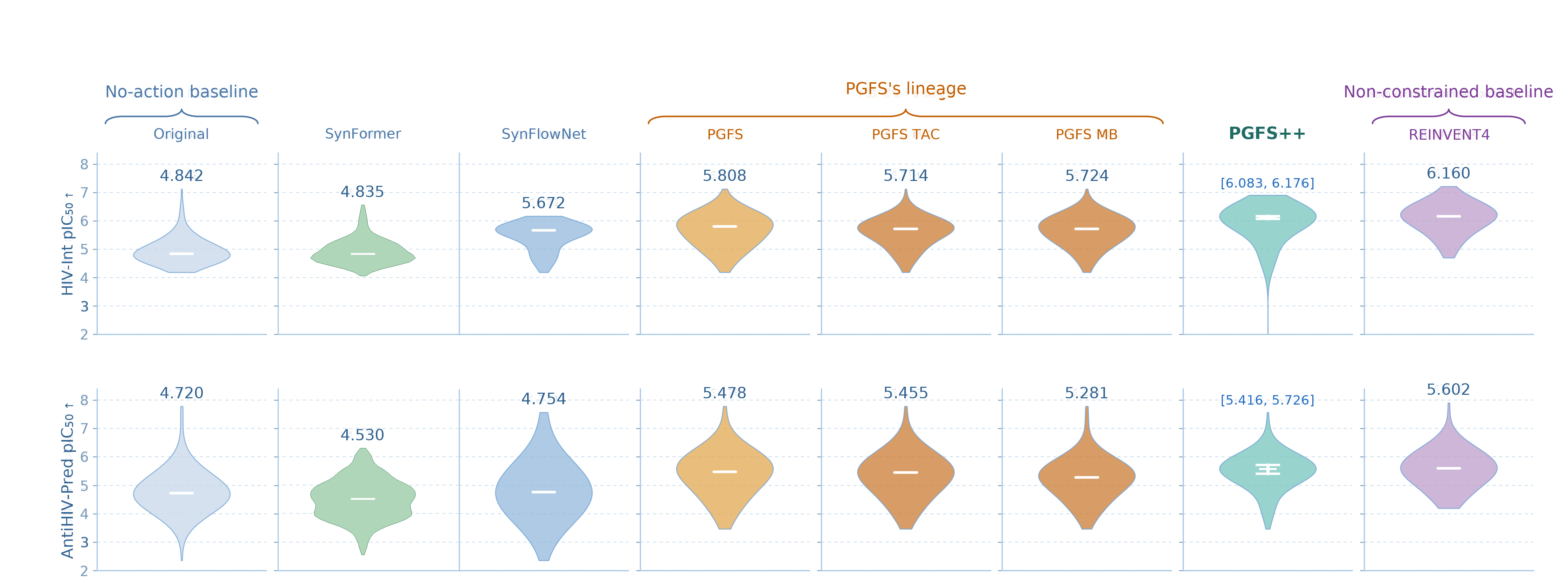}
\caption{Out-of-distribution performance on 200 randomly selected drug-like ChEMBL molecules. AntiHIV-Pred pIC$_{50}$ is an external integrase model for independent evaluation. }
\label{fig:hiv_chembl_ood}
\end{figure*}

To assess whether PGFS++ generalises to molecular inputs beyond the starting building blocks used in the main evaluation, we evaluated the models on 200 randomly selected ChEMBL molecules. These molecules served as starting structures for molecular optimisation, and the resulting products were assessed using the HIV-Int and AntiHIV-Pred integrase predictors. As Figure~\ref{fig:hiv_chembl_ood} shows, PGFS++ achieved higher median predicted potency than the other evaluated synthesis-constrained methods under both predictors. Its median $\mathrm{pIC}_{50}$ increased from 4.842 to 6.129 under HIV-Int and from 4.720 to 5.571 under AntiHIV-Pred, with across-seed standard deviations of 0.046 and 0.155, respectively. These median scores were comparable to those of REINVENT4 (6.160 and 5.602): both differences were 0.031, within one corresponding PGFS++ across-seed standard deviation. PGFS++ thus achieved similar performance while additionally providing explicit synthesis routes. These results indicate that PGFS++ retained its ability to improve predicted potency on ChEMBL inputs beyond the building blocks used in the main evaluation.

\subsubsection{Chemical case studies}

\begin{figure*}[!tp]
\centering
\includegraphics[width=\textwidth]{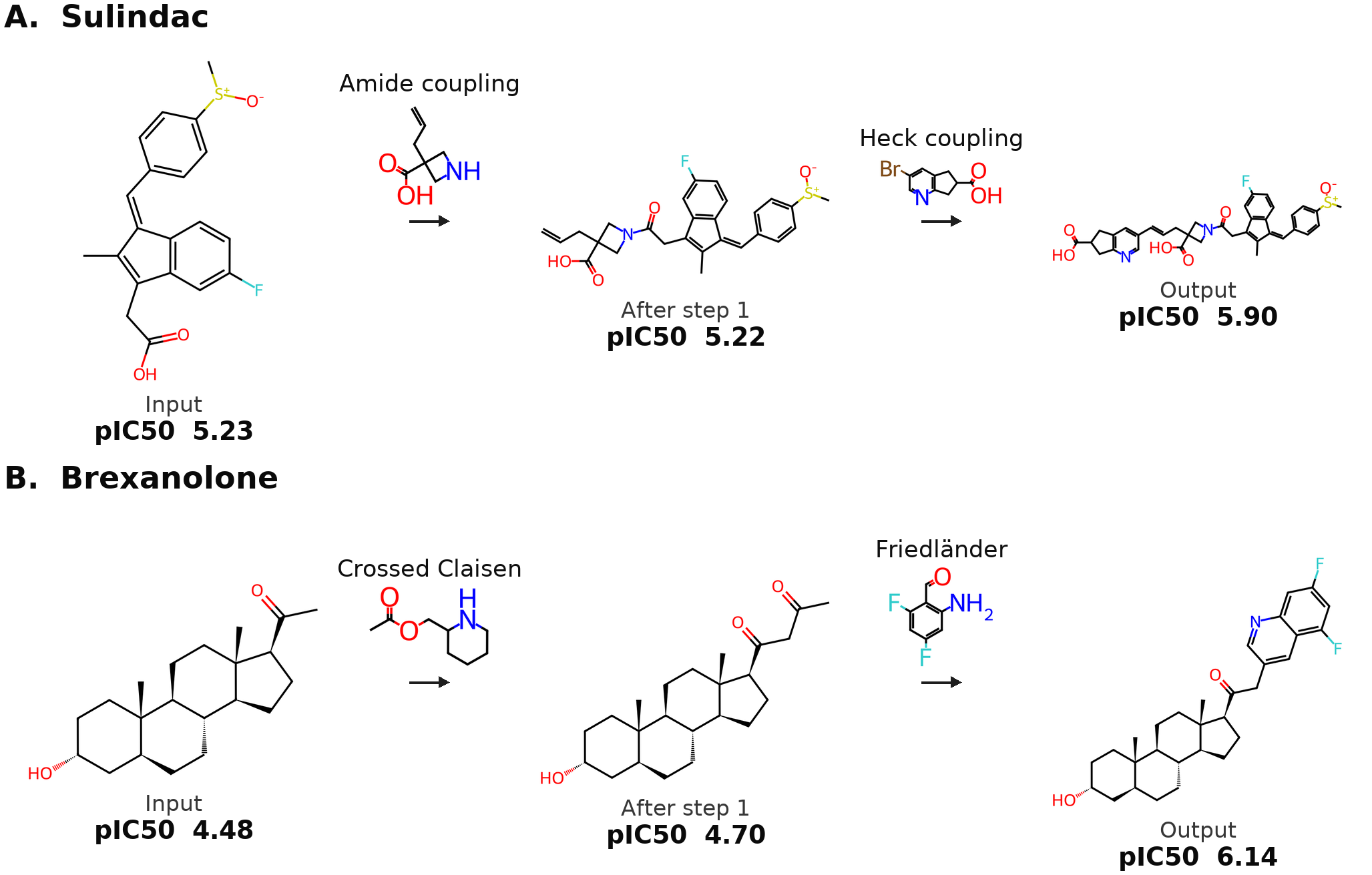}
\caption{Two examples of HIV-Int routes from PGFS++. Each arrow is one template application, with its second reactant shown above the arrow, and the reaction name written above the second reactant.}
\label{fig:hiv_case_routes}
\end{figure*}

Figure~\ref{fig:hiv_case_routes} illustrates how PGFS++ modifies sulindac and brexanolone through proposed reaction routes that increase predicted HIV integrase inhibitory potency. Each route shows the starting molecule, selected reaction templates and second reactants, intermediate products, and predicted $\mathrm{pIC}_{50}$ at each step. All predictions are means from the 25-model ensemble described in Section~\ref{sec:objectives}. 

The sulindac route combines a biologically motivated starting scaffold with a chemically coherent sequence of amidation and side-chain extension. A docking-guided screen of approved drugs identified sulindac as an inhibitor of the HIV-1 integrase--lens epithelium-derived growth factor p75 (LEDGF/p75) interaction~\cite{pgfscase_hu2012}. The first step couples sulindac's carboxylic acid with the nitrogen of 3-allylazetidine-3-carboxylic acid, preserving the fluoroindene and methylsulfinylbenzylidene portions. Related sulindac amides retaining the methylsulfinylbenzylidene group have been synthesised experimentally, providing precedent for this type of modification~\cite{pgfscase_mathew2017}. This amidation leaves predicted $\mathrm{pIC}_{50}$ nearly unchanged, from 5.23 to 5.22, while introducing an allyl group for subsequent functionalisation. The terminal alkene then serves as the reactive handle for a proposed Heck coupling with the heteroaryl bromide, extending the side chain. Heck couplings of bromopyridines with acrylates and styrenes provide related reaction-class precedent, although the present route uses an unactivated allyl substrate~\cite{pgfscase_heck2023}. The second transformation increases predicted $\mathrm{pIC}_{50}$ to 5.90. Together, these steps yield a structurally traceable sulindac derivative through recognised bond-forming transformations, with a modest overall increase in predicted potency.

The brexanolone route combines steroid side-chain functionalisation with quinoline formation, increasing predicted $\mathrm{pIC}_{50}$ from 4.48 to 6.14 while retaining the steroid framework. The first step proposes crossed Claisen condensation with (piperidin-2-yl)methyl acetate to generate a $\beta$-diketone intermediate, increasing the prediction to 4.70. Related Claisen condensations of pregnenolone-derived substrates with dimethyl oxalate have produced steroidal 1,3-dicarbonyl intermediates, providing precedent for this type of side-chain functionalisation~\cite{pgfscase_szabo2015}. The resulting $\beta$-diketone supplies an enolisable carbonyl system for subsequent Friedl\"ander condensation with 2-amino-4,6-difluorobenzaldehyde. This step forms the difluoroquinoline moiety and increases predicted $\mathrm{pIC}_{50}$ to 6.14. Quinoline derivatives of progesterone and pregnenolone have also been synthesised experimentally through a related oxidative annulation, supporting the chemical plausibility of combining these structural elements~\cite{pgfscase_patra2022}. The proposed sequence therefore connects two recognisable synthetic operations: installation of a 1,3-dicarbonyl intermediate and its conversion into a nitrogen-containing heteroaromatic ring. The introduced quinoline additionally connects the product to a scaffold investigated experimentally in allosteric HIV-1 integrase inhibitors~\cite{pgfscase_dinh2022}. Together, these precedents support the route's underlying chemical logic and provide a medicinal chemistry context for the generated product, without assigning it a specific binding mode.

Together, these examples illustrate how PGFS++ generates input-related molecular modifications through chemically interpretable reaction sequences. The sulindac route extends the starting scaffold through amidation and proposed Heck coupling, whereas the brexanolone route introduces a quinoline moiety through side-chain functionalisation and heterocycle formation. The final products retain recognisable features of their starting molecules, with Morgan fingerprint Tanimoto similarities of 0.51 and 0.50, respectively. 

\subsection{Optimzation of other molecular objectives}

\begin{figure*}[!tp]
\centering
\includegraphics[width=\textwidth]{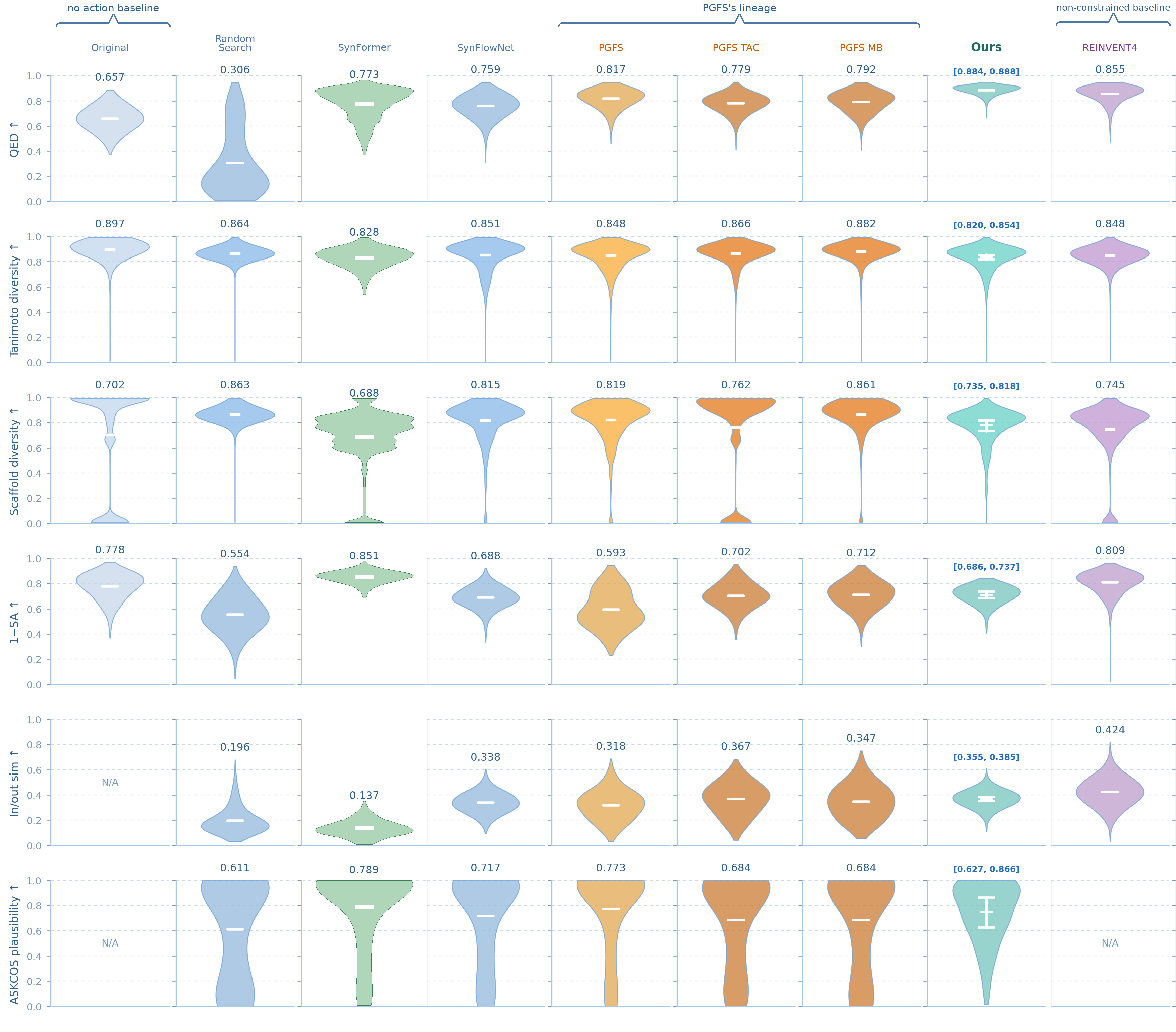}
\caption{Performance comparison on the QED improving task. }
\label{fig:qed_main_experiments}
\end{figure*}

\begin{figure*}[!tp]
\centering
\includegraphics[width=\textwidth]{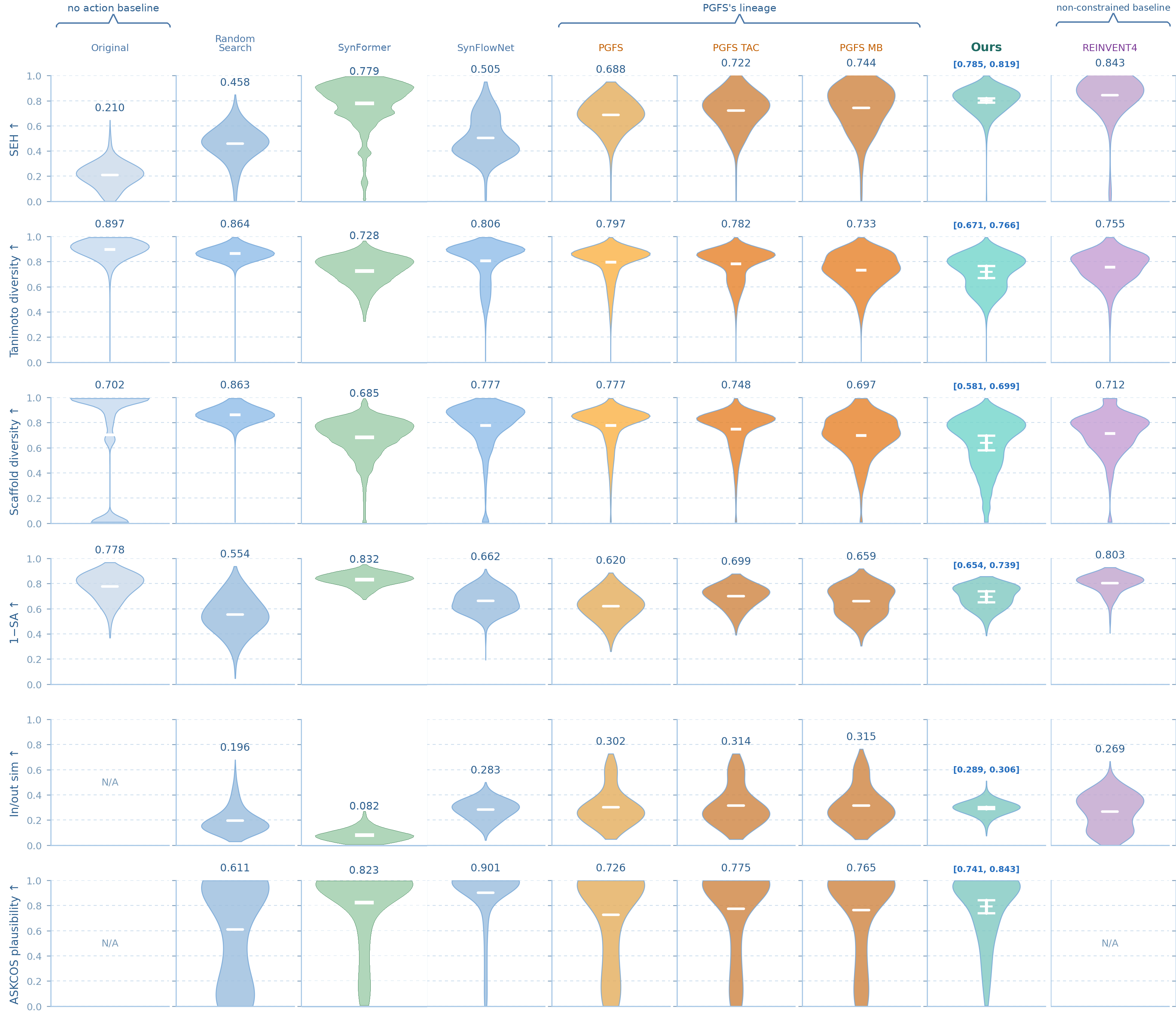}
\caption{Performance comparison on the sEH improving task. }
\label{fig:seh_main_experiments}
\end{figure*}

We further evaluated PGFS++ on two additional optimisation objectives: quantitative estimate of drug-likeness (QED)~\citep{qed} and predicted binding affinity for soluble epoxide hydrolase (sEH)~\citep{seh}. Across both objectives (Figs.~\ref{fig:qed_main_experiments} and~\ref{fig:seh_main_experiments}), PGFS++ consistently outperformed PGFS and its variants while achieving performance comparable to REINVENT4. Unlike REINVENT4, PGFS++ additionally generated explicit synthesis routes for the optimized molecules.
SynFlowNet is included in Figs.~\ref{fig:qed_main_experiments} and~\ref{fig:seh_main_experiments} with the same general templates as the other methods. Its comparison against PGFS++ on reversible templates, using the original QED and sEH scores, is given in the Supplementary Information.

\subsection{Details of route plausibility}
\label{sec:route-plausibility}

Table~\ref{tab:route-plausibility} reports the synthesis routes behind the HIV-Int, QED and sEH comparisons for the 2{,}000 outputs. Route length is the number of reactions per episode. Per-step plausibility is the mean ASKCOS fast-filter score $p_n$, and route plausibility is the mean of $\prod p_n$ over routes that contain at least one reaction. The successful reaction rate is the fraction of attempted reactions that returned a product. REINVENT4 does not provide an explicit synthesis route, so route length, plausibility and the successful reaction rate are not applicable.

\begin{table}[!tp]
\caption{Detailed comparison of ASKCOS plausibility. Entries for PGFS++ are the mean (standard deviation) over three seeds. A dash marks a metric that requires an explicit synthesis route and hence not applicable for REINVET4.}
\label{tab:route-plausibility}
\centering
\scriptsize
\setlength{\tabcolsep}{4pt}
\begin{tabular}{lcccc}
\toprule
Method & Length & $p_n$ & $\prod p_n$ & Successful reaction rate \\
\midrule
\rowcolor{taskrow}\multicolumn{5}{l}{\textbf{HIV-Int}} \\
PGFS & 1.73 & 0.926 & 0.885 & 1.00 \\
PGFS TAC & 2.85 & 0.904 & 0.771 & 1.00 \\
PGFS MB & 1.91 & 0.925 & 0.871 & 1.00 \\
PGFS++ & 2.56 (0.35) & 0.803 (0.078) & 0.547 (0.234) & 1.00 (0.00) \\
SynFormer & 0.73 & 0.718 & 0.704 & 0.708 \\
SynFlowNet & 3.87 & 0.781 & 0.419 & 0.996 \\
REINVENT4 & -- & -- & -- & -- \\
\midrule
\rowcolor{taskrow}\multicolumn{5}{l}{\textbf{QED}} \\
PGFS & 1.50 & 0.845 & 0.773 & 0.9995 \\
PGFS TAC & 1.21 & 0.733 & 0.685 & 1.00 \\
PGFS MB & 1.49 & 0.766 & 0.684 & 1.00 \\
PGFS++ & 1.71 (0.87) & 0.829 (0.019) & 0.747 (0.119) & 1.00 (0.00) \\
SynFormer & 0.99 & 0.797 & 0.789 & 0.886 \\
SynFlowNet & 2.34 & 0.859 & 0.718 & 0.984 \\
REINVENT4 & -- & -- & -- & -- \\
\midrule
\rowcolor{taskrow}\multicolumn{5}{l}{\textbf{sEH}} \\
PGFS & 2.91 & 0.886 & 0.732 & 1.00 \\
PGFS TAC & 2.63 & 0.899 & 0.781 & 1.00 \\
PGFS MB & 2.81 & 0.904 & 0.772 & 1.00 \\
PGFS++ & 2.59 (0.03) & 0.913 (0.024) & 0.792 (0.051) & 1.00 (0.00) \\
SynFormer & 1.16 & 0.846 & 0.823 & 0.873 \\
SynFlowNet & 1.15 & 0.912 & 0.899 & 0.982 \\
REINVENT4 & -- & -- & -- & -- \\
\bottomrule
\end{tabular}
\end{table}

\subsection{Training and inference time}
\label{sec:runtime}

{Table~\ref{tab:runtime}} reports the training and inference times of PGFS and PGFS++ on the QED and sEH tasks, measured on a single NVIDIA A100 80\,GB GPU. Despite scoring all template-compatible $R_2$ candidates globally, PGFS++ is 1.77 times (QED) and 1.52 times (sEH) faster than PGFS in training, and 2.29 times (QED) and 1.43 times (sEH) faster at inference.

\begin{table}[t]
\caption{Training and inference times of PGFS and PGFS++ on the QED and sEH improvement tasks. Inference time is measured on the 2{,}000-molecule test set. PGFS++ training and inference times are the mean (standard deviation) over three seeds.}
\label{tab:runtime}
\footnotesize
\setlength{\tabcolsep}{3pt}
\setlength{\fboxsep}{1pt}
\begin{tabular}{llcccc}
\toprule
& & \multicolumn{2}{c}{Time} & \multicolumn{2}{c}{PGFS/PGFS++} \\
\cmidrule(lr){3-4} \cmidrule(lr){5-6}
Method & Objective & Training & Inference & Training & Inference \\
\midrule
PGFS   & QED & 15\,h\,59\,m & 9\,m\,26\,s  & \multirow{2}{*}{\(1.77\times\ (0.34\times)\)} & \multirow{2}{*}{\(2.29\times\ (0.60\times)\)} \\
PGFS++ & QED & 9\,h\,16\,m (2\,h\,02\,m) & 4\,m\,21\,s (1\,m\,20\,s)  & & \\
\midrule
PGFS   & sEH & 18\,h\,08\,m & 11\,m\,13\,s & \multirow{2}{*}{\(1.52\times\ (0.11\times)\)} & \multirow{2}{*}{\(1.43\times\ (0.003\times)\)} \\
PGFS++ & sEH & 11\,h\,58\,m (51\,m) & 7\,m\,52\,s (1\,s)  & & \\
\botrule
\end{tabular}
\end{table}

\subsection{Ablation Studies}
\label{sec:ablation}

\subsubsection{Strategies of encouraging output diversity} We first compared three representative strategies applied to the terminal episode reward: a hard penalty, where episodes that generate a high-QED molecule (QED$\ge0.8$) already produced more than $B=25$ times receive zero reward~\citep{reinvent4}
; a soft penalty, where the reward is multiplied by $\exp(-n_s/\lambda)$, with $n_s$ the number of previous occurrences of the molecule and $\lambda=25$~\citep{synthemol}
; and an avoid-list penalty, where the reward is reduced by $\lambda s_{\max}$ ($\lambda=0.35$), with $s_{\max}$ the maximum Tanimoto similarity to a list of over 6{,}000 previously identified magnet molecules.

As shown in {Table~\ref{tab:naive_diversity_methods}}, none of the three output-penalisation strategies maintained both property improvement and molecular diversity. To quantify this trade-off, we defined a balanced score as the harmonic mean of $\Delta$QED and diversity:

\[
S_{\mathrm{bal}}
=
\frac{2\,\Delta\mathrm{QED}\,D}
{\Delta\mathrm{QED}+D},
\]

where $D$ denotes diversity. The hard penalty restored diversity to 0.888 but reduced $\Delta$QED to 0.004, resulting in a balanced score of 0.008. The soft and avoid-list penalties retained greater property improvement but achieved diversity values of 0.101 and 0.494, corresponding to balanced scores of 0.141 and 0.313, respectively. In contrast, the additive and multiplicative input--output similarity bonuses maintained both property improvement and diversity, achieving balanced scores of 0.349 and 0.360, respectively. The multiplicative formulation used in PGFS++ therefore achieved the highest balanced score among the evaluated reward designs. 

\begin{table*}[!tp]
\caption{Performance of output-penalisation and input--output similarity reward designs for mitigating magnet-output collapse. The balanced score is the harmonic mean of $\Delta$QED and diversity and rewards methods that maintain both property improvement and molecular diversity. Arrows indicate the preferred direction of each metric. Entries other than the balanced score are the mean, with the 25th, 50th and 75th percentiles in parentheses.}
\label{tab:naive_diversity_methods}
\footnotesize
\setlength{\tabcolsep}{4pt}
\setlength{\fboxsep}{1pt}
\resizebox{\textwidth}{!}{%
\begin{tabular}{lccc}
\toprule
Reward design &
\(\Delta\mathrm{QED}\,\uparrow\) &
Diversity \(\uparrow\) &
Balanced score \(\uparrow\) \\
\midrule
No penalty (PGFS+) & \(\mathbf{0.261}\ (0.194,\ 0.260,\ 0.328)\) & \(0.023\ (0.000,\ 0.000,\ 0.000)\) & \(0.042\) \\
Hard penalty       & \(0.004\ (0.000,\ 0.000,\ 0.000)\) & \(\mathbf{0.888}\ (0.864,\ 0.898,\ 0.928)\) & \(0.008\) \\
Soft penalty       & \(0.231\ (0.164,\ 0.233,\ 0.301)\) & \(0.101\ (0.000,\ 0.000,\ 0.000)\) & \(0.141\) \\
Avoid-list penalty & \(0.229\ (0.152,\ 0.241,\ 0.317)\) & \(0.494\ (0.000,\ 0.600,\ 0.691)\) & \(0.313\) \\
\midrule
Additive bonus & \(0.219\ (0.150,\ 0.222,\ 0.287)\) & \(0.856\ (0.847,\ 0.880,\ 0.906)\) & \(0.349\) \\
Multiplicative bonus (PGFS++) & \(0.229\ (0.165,\ 0.234,\ 0.297)\) & \(0.843\ (0.833,\ 0.873,\ 0.900)\) & \(\mathbf{0.360}\) \\
\bottomrule
\end{tabular}}
\end{table*}

\subsubsection{Different forms of input-output similarity bonus}  
We then compared two forms of input--output similarity bonus:  multiplicative 
(Equation \ref{eq:shaped}) and additive:
\begin{equation}
\label{eqn:additive_multiplicative_forms}
\begin{aligned}
\quad
  & r_{\mathrm{episode}}^{+} = r_{\mathrm{property}}(m_{\mathrm{out}})
    + w\,\mathrm{sim}(m_{\mathrm{in}},m_{\mathrm{out}}),\\
\quad
  & r_{\mathrm{episode}}^{\times} = r_{\mathrm{property}}(m_{\mathrm{out}})
    \big(1 + c\,\mathrm{sim}(m_{\mathrm{in}},m_{\mathrm{out}})\big).
\end{aligned}
\end{equation}
The two forms differ in how strongly they pull the output
toward the input. For the additive form,
$$
\frac{\partial r_{\mathrm{episode}}^{+}}{\partial\,\mathrm{sim}}=w,
$$
while for the multiplicative form,
$$
\frac{\partial r_{\mathrm{episode}}^{\times}}{\partial\,\mathrm{sim}}
=
c\,r_{\mathrm{property}}(m_{\mathrm{out}}).
$$
Thus, the multiplicative bonus becomes stronger for high-property outputs, where magnet-molecule collapse is most likely to be rewarded.

The multiplicative form also provides a more stable way to control the balance
between property optimization and input--output similarity. Because the similarity
term scales the property reward, the bonus acts as a relative adjustment to the
target-property objective. In contrast, the additive form uses an absolute
bonus whose scale must be carefully tuned.

\begin{figure*}[!tp]
\centering
\includegraphics[width=\textwidth]{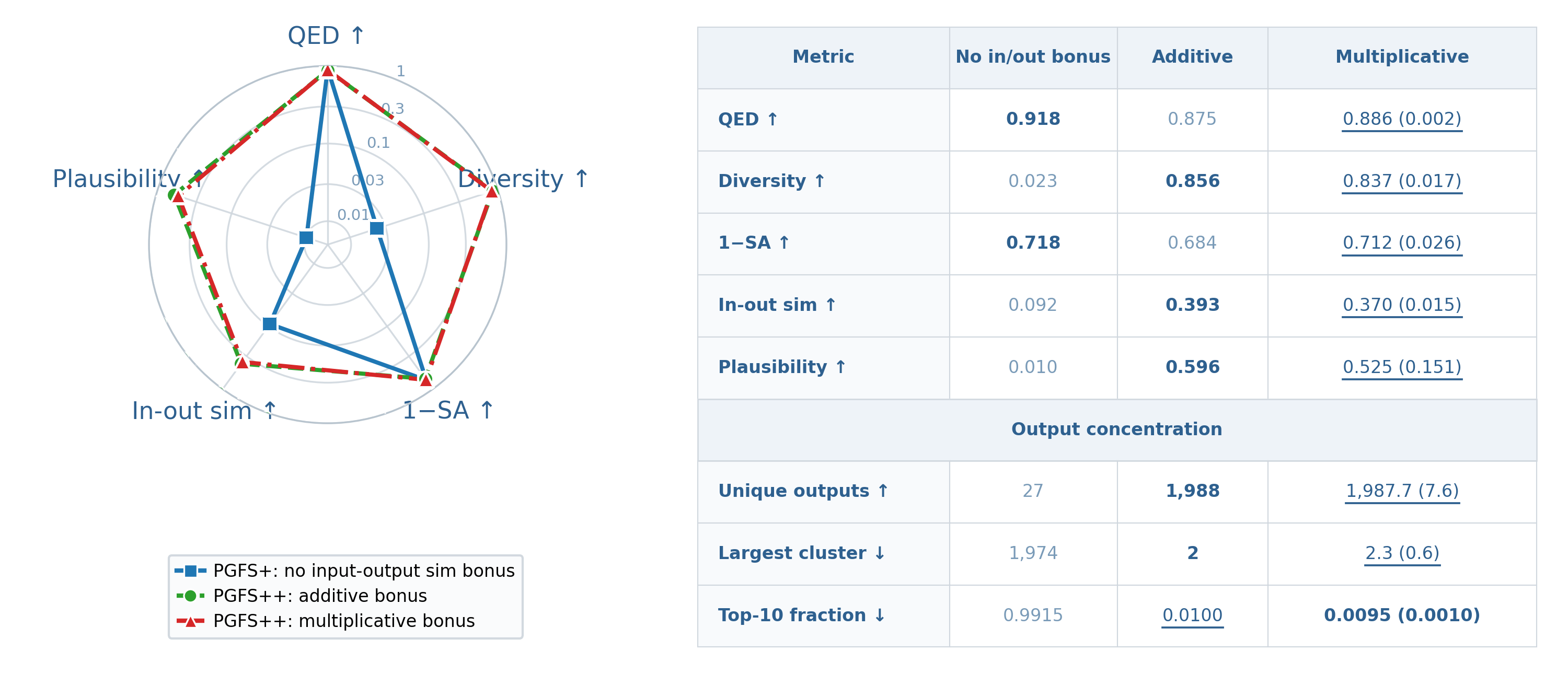}
\caption{Ablation of input--output similarity bonus. We compare PGFS+ without an input--output similarity bonus against PGFS++ with additive and multiplicative similarity bonuses. \textbf{Bold} entries indicate the best value for each metric, and \underline{underlined} entries indicate the second-best value. The multiplicative column is the mean (standard deviation) across three PGFS++ seeds. The output-concentration rows report the number of unique outputs, the size of the largest output cluster, and the fraction of the 2{,}000 inputs mapped to the ten most frequent outputs. Without input--output similarity bonus, PGFS+ suffers from the magnet output problem, as reflected by high mean QED, but very low output diversity, input--output similarity, and ASKCOS plausibility.}
\label{fig:qed_ablation}
\end{figure*}

In Fig.~\ref{fig:qed_ablation}, we compare the two similarity-bonus forms with the
hyperparameters in Equation~\ref{eqn:additive_multiplicative_forms} set to
$w=c=0.5$, while keeping other hyperparameters identical. Both additive and multiplicative similarity bonuses mitigate the magnet
molecule problem observed in PGFS+ without similarity bonus. Compared with PGFS+,
both variants substantially improve output diversity, input--output similarity, and
ASKCOS plausibility while maintaining comparable QED. The two bonus forms achieve similar QED and input--output similarity. Mean diversity and ASKCOS plausibility are higher for the additive form, while mean $1-\mathrm{SA}$ is higher for the multiplicative form.

Without the similarity bonus, 1{,}974 of the 2{,}000 test inputs are mapped to one output. Both bonus forms instead produce nearly 2{,}000 distinct outputs, and about 1\% of inputs fall among the ten most frequent outputs.

\subsubsection{Comparison between similarity bonus and entropy regularization}
We compared the similarity bonus with increasing entropy
regularisation in PGFS+ without the bonus
(Table~\ref{tab:entropy-vs-bonus}).
At an entropy coefficient of $0.01$, the model still exhibited output collapse and produced only
27 unique outputs despite achieving large $\Delta$QED.
Increasing the coefficient alleviated this collapse and
generally increased output diversity, but reduced property
improvement.
At coefficients of $0.05$ and $0.1$, entropy regularisation
achieved diversity comparable to PGFS++, with slightly
smaller QED gains and lower input--output similarity.
Further increases favoured diversity and input similarity
at the expense of QED improvement.
Overall, the empirical results in Table~\ref{tab:entropy-vs-bonus} showed that entropy regularisation can mitigate collapse, but its coefficient requires careful tuning to balance the property improvement, output diversity and input-output similarity.

\begin{table}[t]
\caption{Performance comparison between PGFS++ (multiplicative similarity bonus) and entropy regularization with different coefficients.}
\label{tab:entropy-vs-bonus}
\centering
\footnotesize
\setlength{\tabcolsep}{4pt}
\setlength{\aboverulesep}{0pt}
\setlength{\belowrulesep}{0pt}
\begin{tabular}{l|c|c|c|c}
\toprule
Method & $\Delta$QED & Unique output & Output diversity & Input--output sim. \\
\midrule
PGFS++ & 0.229 (0.002) & 1{,}987.7 (7.6) & 0.837 (0.017) & 0.373 (0.016) \\
\midrule
Entropy $0.01$ & 0.261 & 27 & 0.023 & 0.093 \\
Entropy $0.05$ & 0.215 & 1{,}975 & 0.843 & 0.308 \\
Entropy $0.1$ & 0.217 & 1{,}963 & 0.849 & 0.319 \\
Entropy $0.2$ & 0.132 & 1{,}980 & 0.875 & 0.499 \\
Entropy $0.5$ & 0.047 & 1{,}991 & 0.895 & 0.638 \\
Entropy $1.0$ & 0.029 & 1{,}998 & 0.897 & 0.693 \\
Entropy $2.0$ & 0.014 & 1{,}991 & 0.901 & 0.702 \\
\bottomrule
\end{tabular}
\end{table}

\subsubsection{Direct scoring versus embedding prediction for $R_2$ selection}

To assess the contribution of direct discrete $R_2$ scoring, we constructed an indirect-prediction variant of PGFS++. Following the $R_2$-selection procedure in PGFS, this variant predicted a continuous $R_2$ embedding and used $k$-nearest neighbours ($k$NN) to select a reactant from the available $R_2$ set. To fit with the continuous action in the variant, we also replaced the discrete PPO with a continuous PPO policy, while keeping all other components, including the $T$--$R_2$ compatibility masking and similarity bonus, unchanged. 
As shown in Table~\ref{tab:direct-vs-indirect}, direct scoring achieved a higher $\Delta$QED (0.229 versus 0.217) and output diversity (0.837 versus 0.825). These results show that direct scoring of available $R_2$ candidates is likely a more effective approach over indirect selection through a predicted embedding within the PGFS++ framework.

\begin{table}[t]
\caption{Comparison of $R_2$-selection formulations in PGFS++. Indirect prediction uses a continuous-action PPO policy to predict an embedding, followed by $k$NN-based reactant selection. Direct scoring uses a discrete-action PPO policy to score available candidates. Compatibility masking, the similarity bonus, and all other components are held fixed.}
\label{tab:direct-vs-indirect}
\centering
\footnotesize
\setlength{\tabcolsep}{4pt}
\begin{tabular}{lcc}
\toprule
Method & $\Delta$QED & Output diversity \\
\midrule
PGFS++ (indirect prediction) & 0.217 & 0.825 \\
PGFS++ (direct scoring) & 0.229 (0.002) & 0.837 (0.017) \\
\bottomrule
\end{tabular}
\end{table}

\subsubsection{Comparison of reinforcement learning update rules}

To assess the contribution of PPO~\cite{schulman2017ppo}, we compared it with three representative RL update rules: advantage actor-critic (A2C), a synchronous on-policy baseline following the actor--critic framework of~\cite{mnih2016asynchronous}; advantage-weighted regression (AWR)~\cite{peng2019awr}; and soft actor-critic (SAC)~\cite{haarnoja2018sac,christodoulou2019discrete}, an off-policy method with entropy regularization. These methods provide alternative approaches to policy optimization while retaining the discrete action space.
Previous molecular-generation benchmarks have also reported strong performance for PPO, supporting its use for molecular optimization~\cite{bou2024acegen}. 

\begin{table}[t]
\caption{Performance comparison of RL update rules on the $2{,}000$-molecule test set. PGFS++ is the mean (sample standard deviation) across three seeds. A2C, AWR and SAC are single runs. Reactions per episode is the mean route length.}
\label{tab:rl-backbones}
\centering
\footnotesize
\setlength{\tabcolsep}{4pt}
\begin{tabular}{lccc}
\toprule
Method & $\Delta$QED & Output diversity & Reactions per episode \\
\midrule
PGFS++ (PPO) & 0.229 (0.002) & 0.837 (0.017) & 1.71 (0.87) \\
A2C & 0.189 & 0.839 & 0.95 \\
AWR & 0.039 & 0.879 & 1.13 \\
SAC & $-0.527$ & 0.659 & 3.61 \\
\bottomrule
\end{tabular}
\end{table}

\subsubsection{Effect of $T$-$R_2$ compatibility masking}

To assess the contribution of $T$--$R_2$ compatibility masking, we trained policies without the mask used for $R_2$ selection when choosing a bimolecular reaction template. Without this mask, the policy scored all available $R_2$ candidates, regardless of their compatibility with the selected template. We evaluated this variant both with and without the stop action.

As shown in Table~\ref{tab:compat-masking}, when stopping was allowed, the unmasked policy learned to terminate without performing any reactions, yielding zero QED improvement. This behavior avoided the $-1$ reward assigned to invalid reactions, making immediate termination a safe alternative to reaction selection. The resulting output diversity of 0.897 therefore reflected the unchanged input molecules rather than successful molecular exploration. Disabling the stop action prevented this behavior but reduced $\Delta$QED to $-0.080$ and output diversity to 0.743.
With compatibility masking, PGFS++ achieved a positive $\Delta$QED of 0.229 while maintaining output diversity of 0.837. 

\begin{table}[t]
\caption{Effect of compatibility masking on QED improvement task. PGFS++ is the mean (sample standard deviation) across three seeds. The other rows are single runs. }
\label{tab:compat-masking}
\centering
\footnotesize
\setlength{\tabcolsep}{4pt}
\begin{tabular}{lccc}
\toprule
Method & $\Delta$QED & Output diversity & Reactions per episode \\
\midrule
PGFS++ & 0.229 (0.002) & 0.837 (0.017) & 1.71 (0.87) \\
No mask (stop allowed) & 0.000 & 0.897 & 0.00 \\
No mask (stop not allowed) & $-0.080$ & 0.743 & 3.93 \\
\bottomrule
\end{tabular}
\end{table}

\subsection{Sensitivity to the similarity bonus}
\label{sec:bonus-sensitivity}

The similarity bonus controls the balance between property improvement and retention of input-related structural features through three coefficients: the bonus strength $c$, the similarity threshold $\tau$, and the reward cap $\kappa$ (Eqs.~\ref{eq:shaped} and~\ref{eq:in_out_bonus_reward}). To assess how these coefficients affect this balance on the QED task, we varied one coefficient at a time. The other two were held at their main-experiment values: $c=0.5$, $\tau=0.25$ and $\kappa=0.1$. Table~\ref{tab:bonus-sensitivity} reports the resulting QED improvements, input--output similarities and output diversity.

Increasing the bonus strength $c$ improved input--output similarity with little variation in QED improvement. Across $c\in\{0.25,0.5,1,2\}$, mean similarity increased from 0.367 to 0.401, whereas mean $\Delta$QED remained between 0.228 and 0.236. Within this range, a stronger bonus therefore encouraged greater structural retention while maintaining comparable property gains.

Increasing the similarity threshold $\tau$, however, did not consistently improve structural retention. Mean similarity increased from 0.329 at $\tau=0.1$ to 0.373 at $\tau=0.25$, then decreased to 0.360 and 0.343 at $\tau=0.5$ and $0.75$, respectively. Over the same threshold range, mean $\Delta$QED decreased from 0.234 to 0.207. At the highest threshold, only 0.5\% of outputs attained the required similarity, leaving nearly all episodes subject to the reward cap in Eq.~\ref{eq:in_out_bonus_reward}. Thus, raising the threshold beyond the main-experiment value reduced property improvement without increasing mean similarity.

The reward cap $\kappa$ produced a more consistent trade-off between property improvement and structural retention. Increasing $\kappa$ from 0.01 to 0.3 raised mean $\Delta$QED from 0.230 to 0.242, while reducing mean similarity from 0.377 to 0.329. This reduction was accompanied by a decrease in the proportion of outputs meeting the similarity threshold of 0.25, from 95.6\% to 79.6\%. Across all three coefficient sweeps, output diversity ranged from 0.818 to 0.860.

\begin{table*}[t]
\caption{Sensitivity of PGFS++ to the similarity-bonus coefficients on the QED task. Each policy was trained with seed 246135 for one million environment steps and evaluated on the same 2{,}000 test molecules, allowing at most four reactions per molecule. Entries for $\Delta$QED and input--output similarity report the mean, followed by the 25th, 50th and 75th percentiles in parentheses. The first row gives the main-experiment setting, and subsequent blocks vary one coefficient while holding the other two at this main setting.}
\label{tab:bonus-sensitivity}
\footnotesize
\centering
\begin{tabular}{cccccc}
\toprule
$c$ & $\tau$ & $\kappa$ & $\Delta$QED & Input--output similarity & Diversity \\
\midrule
\multicolumn{6}{l}{\textit{Main-experiment setting}} \\
0.5 & 0.25 & 0.1
& 0.232 (0.162, 0.238, 0.306)
& 0.373 (0.327, 0.375, 0.419)
& 0.846 \\
\midrule
\multicolumn{6}{l}{\textit{Varying bonus strength $c$}} \\
0.25 & 0.25 & 0.1
& 0.232 (0.161, 0.238, 0.307)
& 0.367 (0.311, 0.368, 0.422)
& 0.858 \\
1 & 0.25 & 0.1
& 0.236 (0.173, 0.241, 0.304)
& 0.391 (0.333, 0.390, 0.438)
& 0.842 \\
2 & 0.25 & 0.1
& 0.228 (0.162, 0.234, 0.294)
& 0.401 (0.353, 0.409, 0.450)
& 0.860 \\
\midrule
\multicolumn{6}{l}{\textit{Varying similarity threshold $\tau$}} \\
0.5 & 0.1 & 0.1
& 0.234 (0.162, 0.238, 0.304)
& 0.329 (0.250, 0.333, 0.414)
& 0.818 \\
0.5 & 0.5 & 0.1
& 0.218 (0.159, 0.223, 0.287)
& 0.360 (0.289, 0.360, 0.417)
& 0.845 \\
0.5 & 0.75 & 0.1
& 0.207 (0.152, 0.206, 0.265)
& 0.343 (0.269, 0.340, 0.409)
& 0.843 \\
\midrule
\multicolumn{6}{l}{\textit{Varying reward cap $\kappa$}} \\
0.5 & 0.25 & 0.01
& 0.230 (0.159, 0.235, 0.302)
& 0.377 (0.317, 0.365, 0.417)
& 0.849 \\
0.5 & 0.25 & 0.2
& 0.238 (0.173, 0.242, 0.305)
& 0.362 (0.304, 0.367, 0.417)
& 0.832 \\
0.5 & 0.25 & 0.3
& 0.242 (0.178, 0.250, 0.313)
& 0.329 (0.262, 0.326, 0.390)
& 0.842 \\
\bottomrule
\end{tabular}
\end{table*}

To compare the balance achieved by these policies, we identified the Pareto front with respect to mean $\Delta$QED and mean input--output similarity. A policy is non-dominated if no other tested policy achieves at least as high a value on both metrics and a strictly higher value on one. Four settings were non-dominated: $(c,\tau,\kappa)=(1,0.25,0.1)$, $(2,0.25,0.1)$, $(0.5,0.25,0.2)$ and $(0.5,0.25,0.3)$. These settings spanned mean $\Delta$QED values of 0.228--0.242 and mean similarities of 0.329--0.401. The main-experiment setting achieved 0.232 and 0.373, respectively, and was dominated by the setting with $c=1$, which achieved 0.236 and 0.391. In this comparison, increasing the bonus strength from 0.5 to 1 therefore improved both metrics. Nevertheless, we retained the original hyperparameter settings for the main experiments and did not use this sensitivity analysis to select settings that favoured PGFS++.

The relationship between property improvement and structural retention was also examined within the outputs of the main-experiment policy (Table~\ref{tab:qed-sim-bins}). These outputs were grouped by input--output Tanimoto similarity, calculated using Morgan fingerprints with radius 2 and 1{,}024 bits. Across successive similarity intervals, mean $\Delta$QED decreased from 0.315 in $[0,0.10)$ ($n=4$) to 0.301 in $[0.10,0.25)$ ($n=80$), 0.262 in $[0.25,0.40)$ ($n=1{,}179$), and 0.179 in $[0.40,0.60)$ ($n=731$). The remaining six outputs were identical to their inputs and had $\Delta$QED $=0$. This pattern associates larger property gains with greater structural changes, although the lowest-similarity group contained only four molecules. Together with the coefficient sweeps, these results show how similarity shaping adjusts the balance between QED improvement and retention of input-related structure.

\begin{table}[t]
\caption{QED improvement grouped by input--output similarity for the main experiment model evaluated on 2{,}000 test molecules.The count $n$ gives the number of outputs in each group, with outputs identical to their inputs are reported separately.}
\label{tab:qed-sim-bins}
\footnotesize
\centering
\begin{tabular}{lcc}
\toprule
Similarity & $n$ & Mean $\Delta$QED \\
\midrule
$[0,0.10)$    & 4         & 0.315 \\
$[0.10,0.25)$ & 80        & 0.301 \\
$[0.25,0.40)$ & 1{,}179   & 0.262 \\
$[0.40,0.60)$ & 731       & 0.179 \\
Identical to the input & 6 & 0.000 \\
\bottomrule
\end{tabular}
\end{table}

\section{Conclusion and Discussion}
\label{sec:conclusion_discussion}

In this work, we introduced PGFS++, a synthesis-constrained reinforcement learning framework for molecular improvement. PGFS++ extends PGFS by using a discrete action space, pre-computed $T$--$R_2$ masks, attention-like global scoring for $R_2$ selection, and input--output similarity encouragement. In experiments on QED and sEH improvement tasks, PGFS++ consistently improved over PGFS and its variants and achieved performance comparable to REINVENT4, SynFormer and SynFlowNet. 

However, these findings are based on computational evaluations, and the generated molecules and proposed routes have not been validated experimentally. The predicted properties and route-plausibility scores can help prioritise candidates for laboratory investigation, but these information remain subject to model error.
Experimental synthesis and biological assays are therefore needed to determine whether the predicted improvements translate into experimentally measurable benefits.

Future work could investigate alternative molecular
representations, including the 35-descriptor MolDSet (RLV2) of PGFS \citep{pgfs} and Molecular ACCess System (MACCS) keys \citep{durant2002maccs,rdkit}, or graph-transformer-based representations, in place of the current Morgan fingerprints.
Further evaluation could also incorporate structure-based objectives, such as docking scores, alongside experimental assessment of selected candidates.

%

\backmatter

\bmhead{Supplementary information}

Supplementary information is available as a separate document, covering algorithm details, hyperparameters, analysis of diversity and inout-output similarity, the evaluation protocol, and additional experimental results.



\section*{Declarations}





\subsection*{Data availability}
The datasets and processed resources used in this study are publicly available at \url{https://github.com/bz317/PGFS_plus_plus}.


\subsection*{Code availability}
The source code, configuration files and scripts used to train and evaluate PGFS++ are publicly available at \url{https://github.com/bz317/PGFS_plus_plus}. The repository also provides instructions for reproducing the training and evaluation procedures.

\subsection*{Author contribution}

Boqiao Zhang conducted the experiments and prepared the original manuscript. Godbless Tamaraebi James, Sai Krishna Gottipati and Andrew Fitzgibbon reviewed and revised the manuscript. Andrew Fitzgibbon additionally supervised the work.

\bibliography{references}

\end{document}


\title[Supplementary Information for PGFS++]{Supplementary Materials for PGFS++: Molecular Property Improvement under Synthesis and Diversity Constraints}
\author[1]{\fnm{Boqiao} \sur{Zhang}}
\author[1]{\fnm{Godbless Tamaraebi} \sur{James}}
\author{\fnm{Sai Krishna} \sur{Gottipati}}
\author*[1,2]{\fnm{Andrew} \sur{Fitzgibbon}}\email{awf@graphcore.ai}
\affil*[1]{\orgname{University of Cambridge}, \orgaddress{\country{United Kingdom}}}
\affil[2]{\orgname{Graphcore}, \orgaddress{\country{United Kingdom}}}
\maketitle

\section{Algorithm details}
\label{sec:supp-algorithm}

\subsection{Training paradigm}

Policy Gradient for Forward Synthesis (PGFS++) optimizes a discrete policy over reaction templates and compatible second reactants using proximal policy optimization (PPO). In an episode, the molecular state is the current first reactant $R^{(1)}$. An encoder maps its Morgan fingerprint (FP) to a latent representation $h=f(\mathrm{FP}(R^{(1)}))$. The template head uses this representation to sample a compatible template $T$. A STOP decision terminates the episode without selecting a second reactant. Otherwise, a query network combines $h$ with a learned embedding of $T$ and scores template-compatible second reactants using learned reactant embeddings.

For a bimolecular reaction, let $\mathcal R^{(2)}$ denote the available second reactants and let $\mathcal C(T)\subseteq\mathcal R^{(2)}$ be those compatible with the chosen template. The masked categorical distribution is
\begin{equation}\label{eq:supp-r2-policy}
\pi(R^{(2)}=r\mid R^{(1)},T)
=\frac{\exp(q^\top k_r)}{\sum_{u\in\mathcal C(T)}\exp(q^\top k_u)},\qquad r\in\mathcal C(T),
\end{equation}
and assigns zero probability to incompatible candidates. Here $q$ is the query constructed from the state and template embeddings and $k_r$ is the learned embedding of reactant $r$. The action log-probability stored for PPO is the sum of the conditional categorical log-probabilities,
\begin{equation}\label{eq:supp-logp}
\log\pi(T,R^{(2)}\mid R^{(1)})
=\log\pi(T\mid R^{(1)})+\log\pi(R^{(2)}\mid R^{(1)},T).
\end{equation}
For a STOP or unimolecular action, only the applicable action terms are included. The environment applies the selected reaction and returns the resulting molecule and reward. The reward design is defined in the main text.

The critic estimates the state value $V(\mathrm{FP}(R^{(1)}))$. Training collects on-policy transitions, estimates advantages with generalized advantage estimation (GAE), and updates the actor and critic using the clipped PPO objective. Algorithm~\ref{alg:supp-ppo} summarizes the update procedure.

\begin{algorithm}[t]
\caption{PGFS++ on-policy training procedure}\label{alg:supp-ppo}
\begin{algorithmic}[1]
\State Initialize categorical policy $\pi_\theta$, critic $V_\phi$, and rollout buffer $\mathcal D$
\For{each training iteration}
  \State Empty $\mathcal D$
  \For{$t=1$ to $n_{\mathrm{steps}}$}
    \If{the preceding episode terminated}
      \State Sample a new initial molecule $R_0^{(1)}\sim p_0$
    \EndIf
    \State Sample compatible template $T$ and any required $R^{(2)}$ from $\pi_\theta$
    \State Store the joint action log-probability and $V_\phi(R^{(1)})$
    \State Apply the reaction (or STOP), obtain reward $r_t$, and append transition to $\mathcal D$
  \EndFor
  \State Estimate advantages $\hat A_t$ using GAE and returns $\hat G_t=\hat A_t+V_\phi(R_t^{(1)})$
  \For{$k=1$ to $K_{\mathrm{epoch}}$}
    \For{each minibatch $B\subset\mathcal D$}
      \State Compute $\rho_t=\pi_\theta(a_t\mid R_t^{(1)})/\pi_{\mathrm{old}}(a_t\mid R_t^{(1)})$
      \State Update policy and critic using Eq.~\eqref{eq:supp-ppo} and gradient clipping
    \EndFor
    \If{approximate Kullback--Leibler (KL) divergence exceeds the target}
      \State \textbf{break}
    \EndIf
  \EndFor
\EndFor
\end{algorithmic}
\end{algorithm}

\subsection{PPO objective and hyperparameters}

PGFS uses the off-policy twin delayed deep deterministic policy gradient (TD3) algorithm, whereas PGFS++ uses on-policy PPO to learn categorical decisions over template and reactant indices. For a ratio $\rho_t=\pi_\theta(a_t\mid R_t^{(1)})/\pi_{\mathrm{old}}(a_t\mid R_t^{(1)})$, the loss is
\begin{equation}\label{eq:supp-ppo}
\mathcal L(\theta,\phi)
=-\mathcal L^{\mathrm{clip}}(\rho_t,\hat A_t)
+c_v\mathcal L_V(\phi)-c_e H[\pi_\theta],
\end{equation}
where $\mathcal L^{\mathrm{clip}}$ is the clipped policy surrogate with clipping parameter $\epsilon$, $\mathcal L_V$ is the value-function loss, and $H$ is policy entropy. Advantage estimates use GAE with discount factor $\gamma$ and trace factor $\lambda$. The additional hyperparameters are listed in Table~\ref{tab:supp-ppo-hyper}.

\begin{table}[t]
\caption{PPO training hyperparameters for PGFS++.}\label{tab:supp-ppo-hyper}
\centering
\begin{tabular}{@{}lr lr@{}}
\toprule
Parameter & Value & Parameter & Value \\
\midrule
Learning rate $\alpha$ & $3\times10^{-4}$ & GAE $\lambda$ & $0.95$ \\
Rollout length $n_{\mathrm{steps}}$ & $2048$ & PPO clipping $\epsilon$ & $0.2$ \\
Minibatch size $B$ & $64$ & Entropy coefficient $c_e$ & $0.05$ \\
Update epochs $K_{\mathrm{epoch}}$ & $10$ & Value-loss coefficient $c_v$ & $0.5$ \\
Discount $\gamma$ & $0.99$ & Maximum gradient norm $g_{\max}$ & $0.5$ \\
 & & KL target & $0.02$ \\
\bottomrule
\end{tabular}
\end{table}

\section{Diversity drop and input--output similarity}
\label{sec:supp-diversity}

\subsection{Diversity drop as inward edit bias}
\label{sec:supp-inward-edit}

Molecular optimisation changes both the properties of individual
molecules and output diversity.
To describe these changes, consider $N$ input--output pairs
$(x_i,y_i)$ generated by a policy.
Each molecule is represented by a non-zero binary Morgan fingerprint.
For these fingerprints, Tanimoto similarity defines a positive-semidefinite kernel with unit diagonal
\citep{ralaivola2005graph,tripp2023tanimoto}.
The corresponding feature map $\psi$ therefore satisfies
\[
k(m,n)
=
\mathrm{sim}(m,n)
=
\langle\psi(m),\psi(n)\rangle
\]
Because $k(u,u)=1$, the corresponding feature vectors have
unit norm.
This normalisation applies to the kernel feature representation and does not require Euclidean normalisation of the original
fingerprints.

For these paired molecules, let $I$ be an index sampled
uniformly from $\{1,\ldots,N\}$ and define
\[
X=\psi(x_I),\qquad
Y=\psi(y_I),\qquad
A=Y-X.
\]
The displacement $A$ describes the molecular edit in feature
space, while the shared index preserves the input--output
pairing.
All expectations below refer to this empirical distribution,
conditional on the evaluated policy and outputs.

For a feature-space random variable $Z$, its mean and scalar
variance are
\[
\mu_Z=\mathbb{E}[Z],
\qquad
\operatorname{Var}(Z)
=
\mathbb{E}\!\left[\|Z-\mu_Z\|^2\right].
\]
For two such variables, their scalar covariance is
\[
\operatorname{Cov}(Z,W)
=
\mathbb{E}\!\left[
\langle Z-\mu_Z,W-\mu_W\rangle
\right].
\]
This covariance is an expected inner product of centred
features, rather than a covariance matrix or a normalised
correlation coefficient.
These empirical expectations use the divisor $N$.

The unit-norm representation connects these quantities to
Tanimoto distance, $d(u,v)=1-k(u,v)$.
Specifically, input and output diversity satisfy
\begin{equation}
\label{eq:supp-diversity-variance}
\begin{aligned}
D_{\mathrm{in}}
&=
\frac{1}{N^2}\sum_{i,j}d(x_i,x_j)
=
\operatorname{Var}(X),\\
D_{\mathrm{out}}
&=
\frac{1}{N^2}\sum_{i,j}d(y_i,y_j)
=
\operatorname{Var}(Y).
\end{aligned}
\end{equation}
These averages include all ordered pairs, including
self-comparisons.
The diversity summaries defined in the main text instead
average over distinct pairs $i<j$.
Writing those summaries as $\widetilde D$, the corresponding
all-pairs quantities are
\begin{equation}
\label{eq:supp-diversity-convention}
D_{\mathrm{in}}
=
\frac{N-1}{N}\widetilde D_{\mathrm{in}},
\qquad
D_{\mathrm{out}}
=
\frac{N-1}{N}\widetilde D_{\mathrm{out}}.
\end{equation}

The relation $Y=X+A$ now gives
\begin{equation}
\label{eq:supp-inward-decomposition}
D_{\mathrm{out}}
=
D_{\mathrm{in}}+V_A+2\Gamma,
\qquad
V_A=\operatorname{Var}(A),
\quad
\Gamma=\operatorname{Cov}(X,A).
\end{equation}
Here, $V_A$ measures variation among the edits, whereas
$\Gamma$ measures their alignment with centred input features.
A negative $\Gamma$ indicates that the edits tend to oppose
the directions separating the inputs from their mean.
This inward alignment reduces diversity when
$2\Gamma+V_A<0$.

Complete collapse illustrates the limiting case of this
cancellation.
If every input produces the same molecule $m^\star$, then
$Y$ is constant and
\[
A-\mu_A=-(X-\mu_X).
\]
Consequently,
\[
V_A=D_{\mathrm{in}},
\qquad
\Gamma=-D_{\mathrm{in}},
\qquad
D_{\mathrm{out}}=0.
\]
The edits therefore cancel all centred input variation,
even though the edits themselves vary across inputs.

This decomposition characterises the geometry of collapse.
A property-only objective can favour such collapse when a
common high-scoring molecule is reachable from different inputs.
We next examine conditions under which the similarity-shaped
reward changes this preference and bounds the resulting
diversity loss.

\subsection{Bounding diversity loss through input--output similarity}
\label{sec:supp-similarity-bound}

\paragraph{Assumptions.}
Let $R_i(y)$ denote the shaped episode reward for generating
output molecule $y$ from input molecule $x_i$, and let
$k(x_i,y)$ denote their fingerprint-based similarity.
We set $\tau=a=0.25$ and $b=0.75$, consistent with our
implementation.
To derive a theoretical bound on the loss of output diversity,
we introduce the following assumptions. The first
three characterize the reward structure and the behavior of an idealized policy, while the remaining four specify the chemical
space and sampling distributions.

\begin{enumerate}

\item
\textbf{Existence of rewarding local alternatives.}
For every input $x_i$, there exists at least one reachable
output $y$ satisfying
\[
a\leq k(x_i,y)\leq b,
\qquad
R_i(y)>\kappa.
\]
Thus, each input admits at least one sufficiently rewarding
output within the specified similarity range.

\item
\textbf{Limited reward for highly similar outputs.}
For every reachable output $y$ with $k(x_i,y)>b$,
\[
R_i(y)\leq\kappa.
\]
Hence, outputs exceeding the upper similarity threshold
cannot attain rewards above $\kappa$.

\item
\textbf{Conditional distribution of outputs.}
We assume an idealized policy that selects an output with
$R_i(y)>\kappa$ whenever such an output is available.
Under the input--output similarity reward with $\tau=a$, together with Assumptions 1 and 2, 
every selected output $Y_i$ satisfies
\[
a\leq k(x_i,Y_i)\leq b
\]
and forms an annulus centered for an given input.
\item
\textbf{Chemical space.}
We represent the candidate chemical space by the set
of nonzero binary fingerprints,
\[
\Omega=\{0,1\}^{\ell}\setminus\{\mathbf{0}\},
\]
where $\ell=1024$ in our implementation.

\item
\textbf{Independent and uniformly distributed inputs.}
The input fingerprints $X_1,\ldots,X_N$ are assumed to be
independent and uniformly distributed over $\Omega$,
with a fixed sample size $N$ ($N=2000$ in our test set).

\item
\textbf{Uniform conditional output distribution.}
For each input fingerprint $x\in\Omega$, define the
nonempty set of admissible outputs
\[
\mathcal A(x)
=
\{y\in\Omega:a\leq k(x,y)\leq b\}.
\]
We assume that the conditional distribution of the
generated output is uniform over this set:
\[
Y_i\mid X_i=x
\sim\operatorname{Uniform}(\mathcal A(x)).
\]

\item
\textbf{Equivalent rewards and independent sampling.}
For a given input, all possible outputs in
$\mathcal A(x)$ are assumed to receive the same reward.
Combined with Assumption 5, this implies that the
input--output pairs $(X_i,Y_i)$ are independent.
\end{enumerate}

\noindent Under these assumptions, we first characterize the
marginal distribution of the generated outputs and
subsequently derive a bound on diversity drop, where
diversity is averaged over all $N^2$ ordered pairs.
As a reference, we consider a fully collapsed policy
that maps every input to the same output molecule
$m^\star$ and has zero output diversity.

\paragraph{Distribution of the outputs: close to uniform distribution.}

The conditional sampling assumption specifies how an output
is selected for a given input.
To obtain the output distribution across all inputs, we
average these conditional distributions over the input
distribution.

\noindent Recall that
\[
\Omega_\ell
=
\{0,1\}^{\ell}\setminus\{0\},
\qquad \ell=1024,
\]
is the set of all nonzero binary fingerprints of length
$\ell$, with $|\Omega_\ell|=2^\ell-1$.
The assumed uniform input distribution, denoted by $Q$,
therefore assigns probability
\[
\Pr(X=x)=Q(x)=\frac{1}{|\Omega_\ell|}
\]
to each fingerprint of input $x\in\Omega_\ell$.

\noindent For a given input $x$, the allowed output fingerprints
form the set
\[
\mathcal A(x)
=
\{y\in\Omega_\ell:a\leq k(x,y)\leq b\}.
\]
Uniform sampling from this set gives the conditional
probability
\[
\Pr(Y=y\mid X=x)
=
\frac{\mathbf{1}\{y\in\mathcal A(x)\}}{|\mathcal A(x)|}
\]
where the indicator equals one when $y$ belongs to the input's annulus and zero otherwise.

Averaging this conditional probability over all inputs
gives the marginal output distribution:
\begin{equation}
\label{eq:supp-annulus-output-distribution}
\begin{aligned}
P_Y(y)
&:=\Pr(Y=y)\\
&=\sum_{x\in\Omega_\ell}
\Pr(X=x)\Pr(Y=y\mid X=x)\\
&=
\frac{1}{|\Omega_\ell|}
\sum_{x\in\Omega_\ell}
\frac{\mathbf{1}\{y\in\mathcal A(x)\}}
{|\mathcal A(x)|}.
\end{aligned}
\end{equation}

To quantify their departure from uniformity, consider
an independent reference fingerprint $W\sim Q$.
Unlike the selected output $Y$, this reference fingerprint
is drawn without restricting its similarity to $X$.
The probability that this unrestricted draw falls outside
the annulus is
\begin{equation}
\label{eq:supp-annulus-exclusion}
\varepsilon_{\mathrm{ann}}
=
\Pr_{X,W\sim Q}
\left(k(X,W)\notin[a,b]\right),
\end{equation}
where $X$ and $W$ are independent.

\noindent For a fixed input $x$, the excluded fraction is
\[
1-\frac{|\mathcal A(x)|}{|\Omega_\ell|}.
\]
This fraction is also the total variation distance between
uniform sampling from $\mathcal A(x)$ and uniform sampling
from $\Omega_\ell$, as shown below.
For distributions $P$ and $Q$ on this finite space, total
variation is defined as
\[
\|P-Q\|_{\mathrm{TV}}
=
\frac12\sum_{y\in\Omega_\ell}|P(y)-Q(y)|
\]
that measures the discrepancy between the probabilities
assigned by the two distributions.

Averaging the conditional discrepancies bounds the
discrepancy of the marginal output distribution:
\begin{equation}
\label{eq:supp-annulus-tv}
\begin{aligned}
\|P_Y-Q\|_{\mathrm{TV}}
&\leq
\sum_{x\in\Omega_\ell}
Q(x)\,
\|P_{Y\mid X=x}-Q\|_{\mathrm{TV}}\\
&=
\sum_{x\in\Omega_\ell}
Q(x)
\left(
1-\frac{|\mathcal A(x)|}{|\Omega_\ell|}
\right)\\
&=
\varepsilon_{\mathrm{ann}}.
\end{aligned}
\end{equation}
Thus, excluding only a small fraction of unrestricted
input--output pairs ensures that the marginal output
distribution remains close to the uniform input distribution.

The excluded fraction can be calculated using the bitwise
structure of uniform binary fingerprints.
Before conditioning on nonzero fingerprints, each bit is
independently active with probability $1/2$.
For two independent fingerprints, a bit belongs to their
union with probability $3/4$.
Conditional on belonging to the union, it belongs to their
intersection with probability $1/3$.
Their union size $U$ and intersection size $I$ therefore satisfy
\[
U\sim\operatorname{Binomial}(\ell,3/4),
\qquad
I\mid U=u\sim\operatorname{Binomial}(u,1/3).
\]

For $U>0$, their Tanimoto similarity is $I/U$, which
concentrates around $1/3$ as the fingerprint length increases.
For $\ell=1024$ and $[a,b]=[0.25,0.75]$, evaluating these
binomial tails and conditioning on both fingerprints
being nonzero gives
\[
\varepsilon_{\mathrm{ann}}
\approx2.58\times10^{-7}.
\]
Equation~\eqref{eq:supp-annulus-tv} therefore establishes
that the marginal output distribution is extremely close
to uniform under the stated fingerprint-sampling model.

\paragraph{Diversity preservation.}

The preceding analysis bounds the difference between
the marginal input and output distributions.
To translate this result into a diversity bound, recall
that $Q$ and $P_Y$ denote these distributions and that
\[
\|P_Y-Q\|_{\mathrm{TV}}\leq\varepsilon_{\mathrm{ann}}.
\]
Diversity depends on pairs of molecules rather than
individual molecules.
For two independent draws, the corresponding pair
distributions therefore satisfy
\[
\|P_Y\otimes P_Y-Q\otimes Q\|_{\mathrm{TV}}
\leq
2\|P_Y-Q\|_{\mathrm{TV}}
\leq2\varepsilon_{\mathrm{ann}},
\]
where $\otimes$ denotes the distribution of independent
draws from the two specified distributions.

To translate this distributional bound into a diversity
bound, we consider the Tanimoto distance in $[0,1]$ assigned to
each pair.
For any function with values in this interval, the
difference between its expectations under two distributions
is at most their total variation distance.
Applying this property to independent molecule pairs gives
\[
\begin{aligned}
&\left|
\mathbb E_{Y,Y'\sim P_Y}[d(Y,Y')]
-
\mathbb E_{X,X'\sim Q}[d(X,X')]
\right|\\
&\qquad\leq
\|P_Y\otimes P_Y-Q\otimes Q\|_{\mathrm{TV}}
\leq2\varepsilon_{\mathrm{ann}}.
\end{aligned}
\]
Thus, the closeness of the pair distributions limits
the difference between their expected Tanimoto distances.
The all-pairs diversity averages contain $N(N-1)$
off-diagonal terms and $N$ zero self-distances.
Independence across episodes therefore gives
\[
\begin{aligned}
\mathbb E[D_{\mathrm{in}}]
&=
\left(1-\frac1N\right)
\mathbb E_{X,X'\sim Q}[d(X,X')],\\
\mathbb E[D_{\mathrm{out}}]
&=
\left(1-\frac1N\right)
\mathbb E_{Y,Y'\sim P_Y}[d(Y,Y')].
\end{aligned}
\]
Combining these expressions with the pairwise-distance
bound yields
\begin{equation}
\label{eq:supp-annulus-expected-diversity}
\boxed{
\left|
\mathbb E[D_{\mathrm{out}}]
-
\mathbb E[D_{\mathrm{in}}]
\right|
\leq
2\left(1-\frac1N\right)
\varepsilon_{\mathrm{ann}}.
}
\end{equation}

This expectation bound describes average behaviour across
repeated collections of sampled episodes.
To obtain a guarantee for one finite collection, define
the signed diversity difference
\[
F=D_{\mathrm{in}}-D_{\mathrm{out}}.
\]
Equation~\eqref{eq:supp-annulus-expected-diversity} implies
\[
\mathbb E[F]
\leq
2\left(1-\frac1N\right)\varepsilon_{\mathrm{ann}}
\leq2\varepsilon_{\mathrm{ann}}.
\]

The deviation of $F$ from its expectation can be bounded
using independence across episode pairs $(X_i,Y_i)$.
Replacing one episode pair changes at most $2(N-1)$
terms in each diversity average.
Each term lies in $[0,1]$, so the resulting change in
$F$ is at most
\[
\frac{4(N-1)}{N^2}\leq\frac4N.
\]
The bounded-differences inequality consequently gives
\[
\Pr\!\left(F-\mathbb E[F]\geq t\right)
\leq
\exp\!\left(-\frac{Nt^2}{8}\right).
\]

\noindent
This inequality bounds the probability that $F$
exceeds its expectation by at least $t$.
To make this probability at most a chosen level
$\eta\in(0,1)$, we solve
\[
\exp\!\left(-\frac{Nt^2}{8}\right)=\eta,
\]
which gives
\[
t=\sqrt{\frac{8\log(1/\eta)}{N}}.
\]
Consequently, with probability at least $1-\eta$,
\[
F\leq
\mathbb E[F]
+\sqrt{\frac{8\log(1/\eta)}{N}}.
\]
Using the previously established bound
$\mathbb E[F]\leq2\varepsilon_{\mathrm{ann}}$ then yields
\[
F\leq
2\varepsilon_{\mathrm{ann}}
+\sqrt{\frac{8\log(1/\eta)}{N}}
\]
with the same probability guarantee.
The right-hand side is non-negative, so the same upper
bound applies to the diversity loss
$\Delta_{\mathrm{div}}=[F]_+$:
\begin{equation}
\label{eq:supp-annulus-drop-bound}
\boxed{
\Delta_{\mathrm{div}}
\leq
2\varepsilon_{\mathrm{ann}}
+
\sqrt{\frac{8\log(1/\eta)}{N}}.
}
\end{equation}

\noindent If choosing $\eta=0.05$, we can get a bound that holds with probability at least $95\%$.
For $N=2000$ and
$\varepsilon_{\mathrm{ann}}\approx2.58\times10^{-7}$,
the upper bound of diversity drop is approximately $0.1095$.

\subsection{Empirical evaluation on QED optimisation task}
\label{sec:supp-diversity-empirical}

Table~\ref{tab:supp-qed-diversity-drop} compares the observed
diversity loss of each PGFS++ seed with the theoretical bound
in Equation~\eqref{eq:supp-annulus-drop-bound}.
For $N=2000$ and $\eta=0.05$, this theoretical reference is
$0.1095$ and all three observed losses lie below it.

The theoretical bound is derived with an important assumption that all input--output
similarity lies in $[0.25,0.75]$.
To examine this requirement empirically,
Table~\ref{tab:supp-qed-annulus} partitions each seed's
$2000$ input--output pairs into those inside the interval
and those below or above it.
As the table shows, between $90.5\%$ and $94.4\%$ of the pairs fall inside the interval.
The remaining pairs allow us to assess whether violations
of the interval materially affect the diversity results.
Removing either tail, or both, changes output diversity
by at most $0.0046$.
The corresponding diversity losses, recomputed using the
input diversity of each retained subset, also remain below
the full-test-set reference of $0.1095$.
Thus, the observed preservation of diversity is not driven
by outputs outside the assumed similarity interval.

\begin{table}[t]
\centering
\caption{Observed diversity loss of three PGFS++ seeds
on the $2000$ molecule test set.
Input and output diversity use all $N^2$ ordered pairs,
including self-comparisons.
The theoretical bound reference is
Equation~\eqref{eq:supp-annulus-drop-bound}, evaluated
at $N=2000$ and $\eta=0.05$.}
\label{tab:supp-qed-diversity-drop}
\begin{tabular}{@{}lccc@{}}
\toprule
Seed
& $D_{\mathrm{out}}$
& Observed diversity drop $\Delta_{\mathrm{div}}$
& Theoretical diversity drop  \\
\midrule
123456 & 0.8430 & 0.0540 & 0.1095 \\
654321 & 0.8177 & 0.0793 & 0.1095 \\
246135 & 0.8489 & 0.0481 & 0.1095 \\
\bottomrule
\end{tabular}
\end{table}

\begin{table}[t]
\centering
\footnotesize
\setlength{\tabcolsep}{4pt}
\caption{Sensitivity of output diversity to the
input--output similarity interval.
Here, $s=k(x_i,y_i)$.
The counts partition each seed's $2000$ pairs into the
annulus $[0.25,0.75]$ and its lower and upper tails.
Output diversity is recomputed for the annulus alone,
after retaining the lower tail ($s\leq0.75$),
after retaining the upper tail ($s\geq0.25$),
and for all outputs.}
\label{tab:supp-qed-annulus}
\begin{tabular}{@{}lccc cccc@{}}
\toprule
& \multicolumn{3}{c}{Number of input--output pairs}
& \multicolumn{4}{c}{$D_{\mathrm{out}}$ on retained outputs} \\
\cmidrule(lr){2-4}
\cmidrule(lr){5-8}
Seed
& $s\in[0.25,0.75]$
& $s<0.25$
& $s>0.75$
& Annulus
& $s\leq0.75$
& $s\geq0.25$
& All \\
\midrule
123456
& 1{,}887 & 100 & 13
& 0.8414 & 0.8426 & 0.8419 & 0.8430 \\
654321
& 1{,}810 & 183 & 7
& 0.8131 & 0.8171 & 0.8138 & 0.8177 \\
246135
& 1{,}876 & 103 & 21
& 0.8464 & 0.8483 & 0.8471 & 0.8489 \\
\bottomrule
\end{tabular}
\end{table}

\section{Additional experimental details and results}
\label{sec:supp-experiments}

\subsection{Data sources and preprocessing}

We used two reaction-template libraries, termed general and reversible, together with the building-block collection distributed with PGFS~\citep{pgfs}. The general library contains 102 templates, comprising 15 unimolecular and 87 bimolecular templates derived from~\citep{dingos_dataset}. The reversible library contains the 105 templates used by SynFlowNet, comprising 13 unimolecular and 92 bimolecular templates~\citep{revintro_cretu2025}.

The distinction between these libraries follows the definition of template reversibility adopted by SynFlowNet~\citep{revintro_cretu2025}. Under this definition, a template is reversible if applying it backward to a product recovers the reactants used in the forward step. The reversible library was constructed to satisfy this criterion, whereas the general library contains both reversible and irreversible templates.

The building-block collection consists of Enamine Global Stock compounds distributed with PGFS~\citep{pgfs,Enamine2026}. The PGFS-derived set contained 129{,}104 stereochemically unique building blocks that matched at least one template as the first reactant. From this filtered collection, we randomly selected 2{,}000 molecules as the test set. The remaining compounds formed the training pool and supplied the second-reactant candidates.

\subsection{Model performance and statistical significance}
\label{sec:supp-significance}

Table~\ref{tab:supp-significance} compares mean target-property scores on the shared 2{,}000-molecule test set using general reaction templates. Each comparator score is the mean across its 2{,}000 outputs. For PGFS++, this mean was calculated separately for three training seeds (123456, 654321 and 246135), and the resulting seed-level means were summarised by their mean and sample standard deviation. In seed order, these means were 6.573, 6.555 and 6.617 for human immunodeficiency virus (HIV) integrase (HIV-Int); 0.886, 0.885 and 0.888 for QED; and 0.782, 0.814 and 0.809 for soluble epoxide hydrolase (sEH).

These seed-level means also formed the basis of the statistical comparisons. For each comparator, $\Delta$ denotes the PGFS++ mean across seeds minus the comparator score. A two-sided one-sample $t$-test assessed whether the three PGFS++ seed-level means differed from that score. Each test therefore treats the comparator score as fixed and accounts only for variation across PGFS++ seeds, rather than training variability in both methods.

PGFS++ achieved higher mean scores than all three comparators on HIV-Int and QED, with statistically significant differences under these tests ($p\leq 0.004$). On sEH, the results were more mixed. REINVENT4 achieved a higher mean score than PGFS++ (0.843 versus 0.802), but the difference did not reach the 0.05 significance threshold ($p=0.052$). PGFS++ scored higher than SynFormer (0.802 versus 0.779), although this difference was also not statistically significant ($p=0.146$). Its advantage over SynFlowNet was larger (0.802 versus 0.505) and statistically significant ($p=0.001$).

\begin{table}[t]
\centering
\renewcommand{\arraystretch}{1.18}
\setlength{\tabcolsep}{5.5pt}
\caption{Mean target-property scores on the $2{,}000$ test molecules using general template set.
HIV-Int is the ensemble predicted $\mathrm{pIC}_{50}$, QED is the RDKit drug-likeness score, and sEH is the scaled soluble-epoxide-hydrolase predictor used in the main figures.
PGFS++ is the mean (sample standard deviation) of three seeds, and the bracket is the 95\% $t$ confidence interval (CI) for that mean.
Comparators are a single run and have no interval.
$\Delta$ is the PGFS++ mean minus the comparator.
The $p$-value is a two-sided one-sample $t$-test of the three seed-level means against the comparator.
Bold marks the higher of the comparator and the PGFS++ mean.
Shaded cells are the PGFS++ summary.}
\label{tab:supp-significance}
\begin{tabular}{@{}llccccc@{}}
\toprule
Objective & Method & Score & PGFS++ & 95\% CI & $\Delta$ & $p$ \\
\midrule
\multirow{3}{*}{HIV-Int}
 & REINVENT4  & $6.278$ & \cellcolor{meanhl}$\mathbf{6.582}\ (0.032)$ & \cellcolor{meanhl}$[6.503,\ 6.660]$ & $+0.304$ & $0.004$ \\
 & SynFormer  & $4.780$ & \cellcolor{meanhl}$\mathbf{6.582}\ (0.032)$ & \cellcolor{meanhl}$[6.503,\ 6.660]$ & $+1.802$ & $<0.001$ \\
 & SynFlowNet & $5.783$ & \cellcolor{meanhl}$\mathbf{6.582}\ (0.032)$ & \cellcolor{meanhl}$[6.503,\ 6.660]$ & $+0.799$ & $<0.001$ \\
\addlinespace[2pt]
\midrule
\addlinespace[1pt]
\multirow{3}{*}{QED}
 & REINVENT4  & $0.855$ & \cellcolor{meanhl}$\mathbf{0.886}\ (0.002)$ & \cellcolor{meanhl}$[0.882,\ 0.890]$ & $+0.031$ & $<0.001$ \\
 & SynFormer  & $0.773$ & \cellcolor{meanhl}$\mathbf{0.886}\ (0.002)$ & \cellcolor{meanhl}$[0.882,\ 0.890]$ & $+0.113$ & $<0.001$ \\
 & SynFlowNet & $0.759$ & \cellcolor{meanhl}$\mathbf{0.886}\ (0.002)$ & \cellcolor{meanhl}$[0.882,\ 0.890]$ & $+0.127$ & $<0.001$ \\
\addlinespace[2pt]
\midrule
\addlinespace[1pt]
\multirow{3}{*}{sEH}
 & REINVENT4  & $\mathbf{0.843}$ & \cellcolor{meanhl}$0.802\ (0.017)$ & \cellcolor{meanhl}$[0.760,\ 0.844]$ & $-0.041$ & $0.052$ \\
 & SynFormer  & $0.779$ & \cellcolor{meanhl}$\mathbf{0.802}\ (0.017)$ & \cellcolor{meanhl}$[0.760,\ 0.844]$ & $+0.023$ & $0.146$ \\
 & SynFlowNet & $0.505$ & \cellcolor{meanhl}$\mathbf{0.802}\ (0.017)$ & \cellcolor{meanhl}$[0.760,\ 0.844]$ & $+0.297$ & $0.001$ \\
\bottomrule
\end{tabular}
\end{table}

\subsection{SynFlowNet performance under general and reversible templates}

The choice of reaction templates affected the QED performance of SynFlowNet and PGFS++ differently (Table~\ref{tab:supp-reversible}). With all other settings held fixed, switching from general to reversible templates increased SynFlowNet's mean QED from 0.759 to 0.838. The same change reduced the PGFS++ mean from 0.886 to 0.858, with an across-seed standard deviation of 0.002 in both settings. Reversible templates therefore narrowed the performance gap between the two methods, although PGFS++ retained a higher mean QED.

\begin{figure}[t]
\centering
\includegraphics[width=\textwidth]{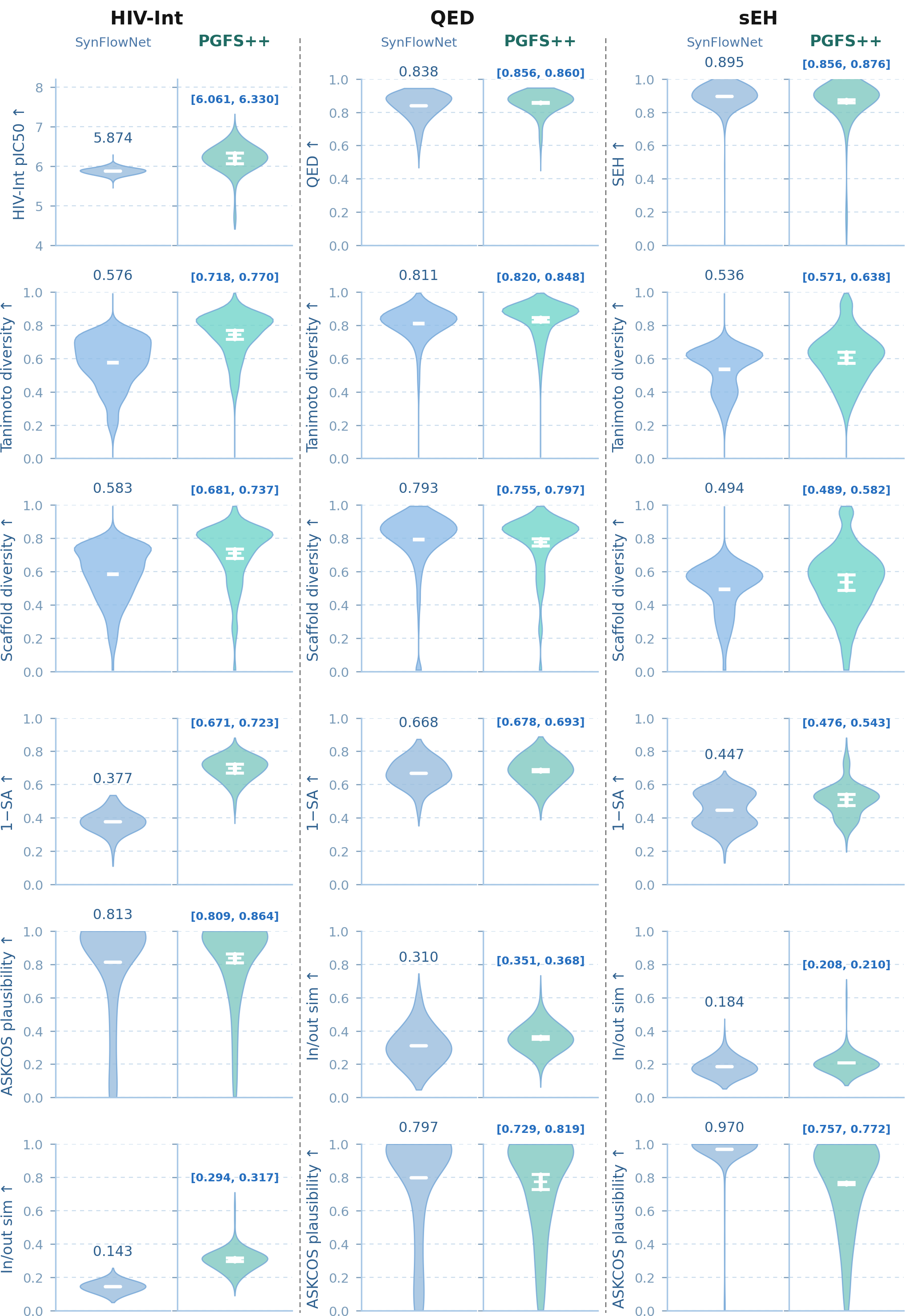}
\caption{Performance comparison of SynFlowNet and PGFS++ using reversible templates. }
\label{fig:supp-reversible-templates}
\end{figure}

To examine their relative performance under reversible templates more broadly, we compared both methods across HIV-Int, QED and sEH (Fig.~\ref{fig:supp-reversible-templates}). This comparison complements the general-template experiments in the main text and includes the same diversity, synthetic-accessibility, input--output similarity and Automated System for Knowledge-based Continuous Organic Synthesis (ASKCOS) plausibility metrics.

Under reversible templates, the relative performance of the two methods varied by objective. On HIV-Int, PGFS++ achieved seed-wise mean predicted $\mathrm{pIC}_{50}$ values of 6.06--6.33, compared with 5.87 for SynFlowNet. On QED, PGFS++ achieved seed-wise means of 0.856--0.860, compared with 0.838 for SynFlowNet. On sEH, SynFlowNet achieved the higher mean score of 0.895, compared with 0.856--0.876 for PGFS++. Thus, PGFS++ remained competitive under reversible templates, achieving higher mean scores on HIV-Int and QED, while SynFlowNet performed better on sEH.

\begin{table}[t]
\caption{Mean QED on the $2{,}000$-molecule test set with general and reversible reaction templates. SynFlowNet is one run. PGFS++ is the mean (standard deviation) across the three seeds. }\label{tab:supp-reversible}
\centering
\begin{tabular}{@{}lcc@{}}
\toprule
Method & General templates & Reversible templates \\
\midrule
SynFlowNet & $0.759$ & $0.838$ \\
PGFS++ & $0.886\ (0.002)$ & $0.858\ (0.002)$ \\
\bottomrule
\end{tabular}
\end{table}

\subsection{Seed variability of PGFS++}

Seed-level performance indicates the stability of the molecular property gains and the variability of other characteristics, especially route plausibility. Table~\ref{tab:supp-seeds} reports the mean of each metric across generated products for each of the three random seeds, followed by the standard deviation of the seed-wise means. 

\begin{table*}[t]
\caption{Seed-wise mean PGFS++ performance on HIV-Int, QED and sEH objectives. Gen. and Rev. indicate general and reversible template sets. SA denotes synthetic accessibility, I/O sim.\ is input--output similarity, and the final column is ASKCOS plausibility. The last two rows within each setting summarize the three seed-wise means.}\label{tab:supp-seeds}
\centering
\scriptsize
\renewcommand{\arraystretch}{1.18}
\setlength{\tabcolsep}{5pt}
\resizebox{\textwidth}{!}{%
\begin{tabular}{|c|c|c|c|c|c|c|c|}
\hline
Templates & Objective & Seed/statistic & Property $\uparrow$ & Diversity $\uparrow$ & $1-\mathrm{SA}$ $\uparrow$ & I/O sim. $\uparrow$ & ASKCOS plausibility $\uparrow$ \\
\hline
Gen. & HIV-Int & 123456 & 6.573 & .741 & .631 & .280 & .285 \\
 & & 654321 & 6.555 & .741 & .640 & .278 & .622 \\
 & & 246135 & 6.617 & .704 & .618 & .302 & .734 \\
 & & Mean & 6.582 & .729 & .630 & .287 & .547 \\
 & & s.d. & .032 & .022 & .011 & .013 & .234 \\
\hline
Gen. & QED & 123456 & .886 & .843 & .738 & .386 & .829 \\
 & & 654321 & .885 & .818 & .687 & .356 & .609 \\
 & & 246135 & .888 & .849 & .710 & .367 & .800 \\
 & & Mean & .886 & .837 & .712 & .370 & .746 \\
 & & s.d. & .002 & .017 & .026 & .015 & .119 \\
\hline
Gen. & sEH & 123456 & .782 & .753 & .728 & .305 & .801 \\
 & & 654321 & .814 & .738 & .648 & .298 & .737 \\
 & & 246135 & .809 & .664 & .713 & .288 & .838 \\
 & & Mean & .802 & .719 & .696 & .297 & .792 \\
 & & s.d. & .017 & .048 & .043 & .009 & .051 \\
\hline
Rev. & HIV-Int & 123456 & 6.193 & .742 & .715 & .318 & .811 \\
 & & 654321 & 6.063 & .771 & .668 & .302 & .835 \\
 & & 246135 & 6.331 & .719 & .708 & .296 & .865 \\
 & & Mean & 6.196 & .744 & .697 & .306 & .837 \\
 & & s.d. & .134 & .026 & .026 & .011 & .027 \\
\hline
Rev. & QED & 123456 & .856 & .839 & .683 & .366 & .806 \\
 & & 654321 & .859 & .819 & .680 & .350 & .793 \\
 & & 246135 & .860 & .846 & .694 & .363 & .722 \\
 & & Mean & .858 & .834 & .685 & .360 & .774 \\
 & & s.d. & .002 & .014 & .007 & .008 & .045 \\
\hline
Rev. & sEH & 123456 & .876 & .628 & .547 & .210 & .758 \\
 & & 654321 & .856 & .621 & .485 & .208 & .772 \\
 & & 246135 & .868 & .566 & .496 & .209 & .764 \\
 & & Mean & .866 & .605 & .509 & .209 & .765 \\
 & & s.d. & .010 & .034 & .033 & .001 & .007 \\
\hline
\end{tabular}}
\end{table*}

\subsection{Novelty of generated molecules}

Molecular novelty was assessed against Molecular Sets (MOSES) \citep{moses}, a collection of approximately 1.5 million molecules. For a generated molecule $m$, novelty is defined as one minus its maximum fingerprint similarity to the reference collection:
\begin{equation}\label{eq:supp-novelty}
\operatorname{novelty}(m)=1-\max_{m'\in\mathrm{MOSES}}
\operatorname{Tanimoto}\bigl(\mathrm{FP}(m),\mathrm{FP}(m')\bigr).
\end{equation}
This measure complements output diversity by testing distance from an external reference set rather than variation among generated molecules. Table~\ref{tab:supp-novelty} compares the mean of the three PGFS++ seed-wise medians with the median of the appropriate single-run baseline. Positive differences indicate higher values for PGFS++.

\begin{table*}[t]
\caption{Target-property and novelty comparison for PGFS++ and the corresponding baseline on the test set. Baseline is REINVENT4 for general templates and SynFlowNet for reversible templates. PGFS++ values are means of three seed-wise medians, whereas each baseline value is the median of a single run. Reported $p$ values reflect paired comparisons over starting molecules.}\label{tab:supp-novelty}
\centering\scriptsize
\resizebox{\textwidth}{!}{%
\begin{tabular}{@{}lllrrrrrrrr@{}}
\toprule
&&&\multicolumn{4}{c}{Median target property}&\multicolumn{4}{c}{Median novelty}\\
\cmidrule(lr){4-7}\cmidrule(lr){8-11}
Templates & Objective & Baseline & Baseline & PGFS++ & $\Delta$ & $p$ & Baseline & PGFS++ & $\Delta$ & $p$ \\
\midrule
General & QED & REINVENT4 & .871 & .896 & +.025 & $<.001$ & .457 & .572 & +.115 & .030 \\
General & sEH & REINVENT4 & .869 & .826 & -.043 & .026 & .426 & .600 & +.174 & .002 \\
Reversible & QED & SynFlowNet & .856 & .869 & +.013 & .031 & .564 & .565 & +.001 & .879 \\
Reversible & sEH & SynFlowNet & .900 & .902 & +.002 & .855 & .645 & .673 & +.028 & .036 \\
\bottomrule
\end{tabular}}
\end{table*}


\subsection{Applicability domain of the HIV-integrase objective}
\label{sec:supp-hiv-ad}
    
The HIV-integrase experiments evaluated each terminal molecule using the 25-model LightGBM ensemble released with PGFS~\citep{pgfs}. To assess whether the resulting predictions fall within the ensemble's applicability domain (AD), we report PGFS++ performance both before and after excluding out-of-domain compounds, following the evaluation in PGFS.

The applicability domain was defined using the distance-based criterion in Appendix~C of PGFS. This criterion uses the 24 RDKit descriptors retained by all models in the ensemble, with each descriptor standardised using the curated ChEMBL training set of 775 compounds. In this standardised feature space, each compound's distance to the training set was calculated as its mean Euclidean distance to the $K=\lfloor\sqrt{775}\rfloor=27$ nearest training neighbours. A compound was considered within the domain when this distance did not exceed the training-set mean of the corresponding distances plus 1.5 standard deviations. This cutoff included 94.1\% of the training compounds.

Using this domain definition, we evaluated each of the three PGFS++ seeds greedily on the same 2{,}000 test reactants, allowing at most four reactions and an explicit STOP action. Tables~\ref{tab:supp-hiv-ad-gen} and~\ref{tab:supp-hiv-ad-rev} report terminal-molecule statistics before and after applicability-domain filtering. For each column, mean predicted $\mathrm{pIC}_{50}$ and mean $\Delta\mathrm{pIC}_{50}$ were calculated over the retained episodes. The top-100 statistic was calculated as the mean predicted $\mathrm{pIC}_{50}$ of the 100 highest-scoring compounds after removing stereochemical annotations, following Table~3 of PGFS.

Filtering affected the top-100 and full-set summaries differently. For the top-100 subset, excluding high-scoring out-of-domain compounds required replacing them with lower-scoring compounds within the domain, reducing the mean. This pattern was also reported in PGFS, where the top-100 mean decreased from $7.16\pm0.09$ to $7.15\pm0.08$ after filtering. The PGFS++ top-100 means decreased in the same direction. Across all terminal molecules, however, the mean increased slightly because the excluded compounds had lower average predicted scores than those retained.

\begin{table*}[t]
\caption{HIV-integrase scores of PGFS++ with general reaction templates, before and after applicability-domain filtering. }\label{tab:supp-hiv-ad-gen}
\centering\small
\begin{tabular}{@{}llcc@{}}
\toprule
Metric & Seed & Without AD filtering & With AD filtering \\
\midrule
\multirow{3}{*}{Mean pIC$_{50}$}
 & 123456 & $6.573\pm 0.348$ & $6.597\pm 0.347$ \\
 & 654321 & $6.555\pm 0.292$ & $6.576\pm 0.296$ \\
 & 246135 & $6.617\pm 0.336$ & $6.631\pm 0.335$ \\
\midrule
\multirow{3}{*}{Mean $\Delta$pIC$_{50}$}
 & 123456 & $1.874\pm 0.421$ & $1.903\pm 0.416$ \\
 & 654321 & $1.856\pm 0.372$ & $1.883\pm 0.372$ \\
 & 246135 & $1.918\pm 0.402$ & $1.935\pm 0.401$ \\
\midrule
\multirow{3}{*}{Top 100}
 & 123456 & $7.243\pm 0.097$ & $7.241\pm 0.099$ \\
 & 654321 & $7.149\pm 0.094$ & $7.148\pm 0.095$ \\
 & 246135 & $7.317\pm 0.088$ & $7.317\pm 0.089$ \\
\midrule
\multirow{3}{*}{Molecules}
 & 123456 & $2{,}000$ & $1{,}652$ ($82.6\%$) \\
 & 654321 & $2{,}000$ & $1{,}644$ ($82.2\%$) \\
 & 246135 & $2{,}000$ & $1{,}890$ ($94.5\%$) \\
\bottomrule
\end{tabular}
\end{table*}

\begin{table*}[t]
\caption{HIV-integrase scores of PGFS++ with the reversible reaction templates, before and after applicability-domain filtering. The evaluation protocol and the applicability-domain criterion are the same as in Table~\ref{tab:supp-hiv-ad-gen}.}\label{tab:supp-hiv-ad-rev}
\centering\small
\begin{tabular}{@{}llcc@{}}
\toprule
Metric & Seed & Without AD filtering & With AD filtering \\
\midrule
\multirow{3}{*}{Mean pIC$_{50}$}
 & 123456 & $6.193\pm 0.386$ & $6.236\pm 0.368$ \\
 & 654321 & $6.063\pm 0.357$ & $6.107\pm 0.326$ \\
 & 246135 & $6.331\pm 0.486$ & $6.354\pm 0.452$ \\
\midrule
\multirow{3}{*}{Mean $\Delta$pIC$_{50}$}
 & 123456 & $1.494\pm 0.443$ & $1.559\pm 0.408$ \\
 & 654321 & $1.364\pm 0.394$ & $1.421\pm 0.354$ \\
 & 246135 & $1.632\pm 0.496$ & $1.657\pm 0.458$ \\
\midrule
\multirow{3}{*}{Top 100}
 & 123456 & $6.830\pm 0.113$ & $6.827\pm 0.115$ \\
 & 654321 & $6.708\pm 0.129$ & $6.708\pm 0.129$ \\
 & 246135 & $7.216\pm 0.115$ & $7.208\pm 0.115$ \\
\midrule
\multirow{3}{*}{Molecules}
 & 123456 & $2{,}000$ & $1{,}586$ ($79.3\%$) \\
 & 654321 & $2{,}000$ & $1{,}714$ ($85.7\%$) \\
 & 246135 & $2{,}000$ & $1{,}857$ ($92.9\%$) \\
\bottomrule
\end{tabular}
\end{table*}

\subsection{Molecular weights and synthesis accessibility}

\begin{figure}[H]
\centering
\includegraphics[width=\textwidth]{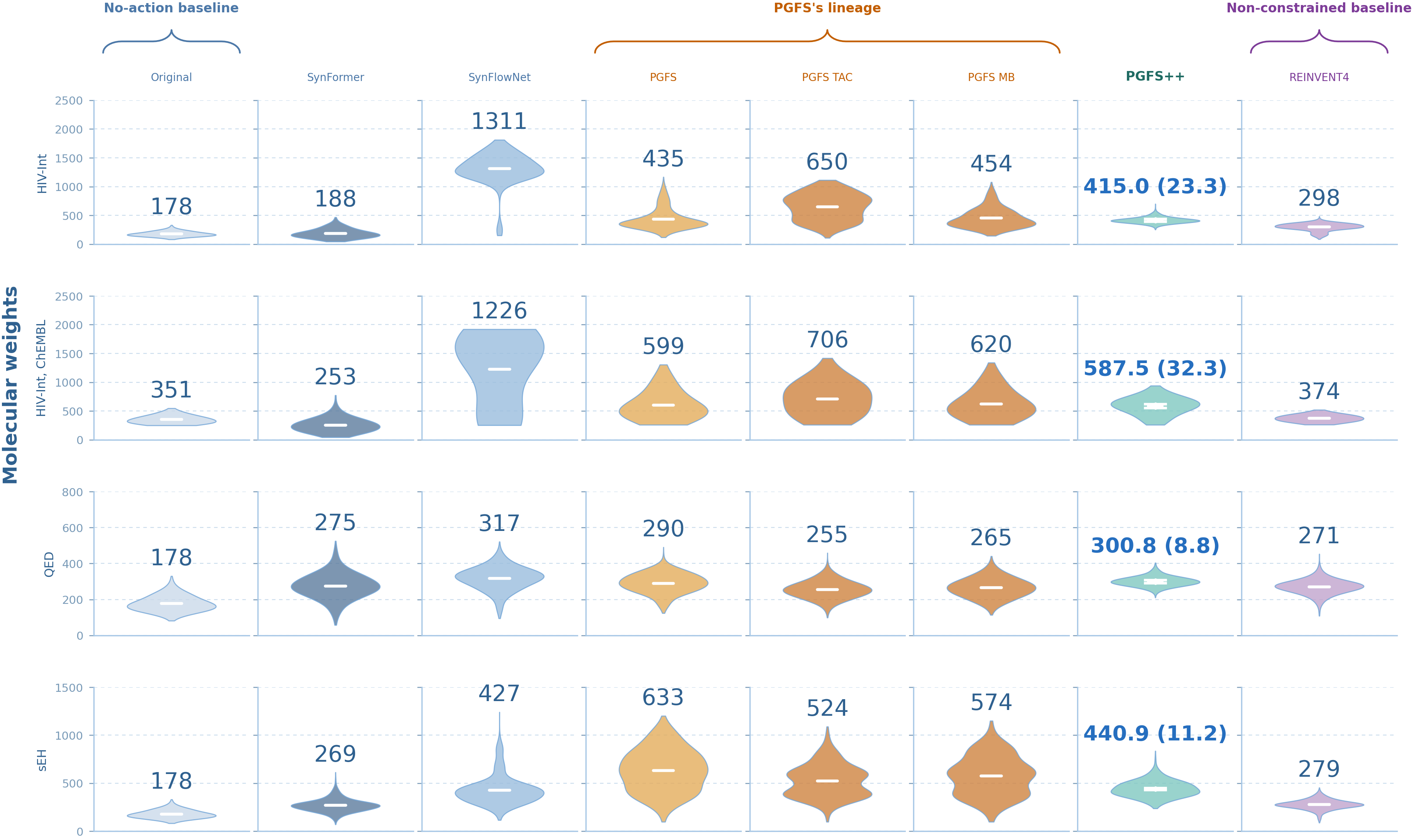}
\caption{Molecular weights of the outputs generated by different methods. For PGFS++, the violin is the per-input mean across the three seeds $\pm$ standard deviation. For other methods, the printed value is the mean.}
\label{fig:supp-mw}
\end{figure}

\makeatletter
\begin{center}
\begingroup
\def\@captype{table}\caption{Molecular weight (MW) and synthetic accessibility (SA) of the outputs generated by different methods.
For PGFS++, MW and SA are the mean (sample standard deviation) of the three seed-level means. For other methods, MW and SA are displayed as the mean values.
Correlation is the Spearman correlation between MW and SA, with each output molecule as one point, and $p$ is the two-sided $p$-value of that correlation.
}
\label{tab:supp-mw-sa}
\endgroup
\renewcommand{\arraystretch}{1.02}
\setlength{\tabcolsep}{2.4pt}
\fontsize{6}{7}\selectfont
\begin{tabular}{@{}llccccccc@{}}
\toprule
Task & & SynFormer & SynFlowNet & PGFS & PGFS TAC & PGFS MB & PGFS++ & REINVENT4 \\
\midrule
\multirow{4}{*}{HIV-Int} & MW $\downarrow$ & $187.5$ & $1311.0$ & $434.8$ & $650.3$ & $454.5$ & $415.0\,(23.3)$ & $298.0$ \\
 & SA $\downarrow$ & $2.30$ & $6.73$ & $3.27$ & $3.91$ & $3.36$ & $3.59\,(0.08)$ & $2.58$ \\
 & Corr & $-0.10$ & $0.83$ & $0.59$ & $0.87$ & $0.73$ & $0.20$ & $0.61$ \\
 & $p$ & $<0.001$ & $<0.001$ & $<0.001$ & $<0.001$ & $<0.001$ & $<0.001$ & $<0.001$ \\
\midrule
\multirow{4}{*}{\shortstack[l]{HIV-Int\\ChEMBL}} & MW $\downarrow$ & $253.3$ & $1225.6$ & $598.7$ & $706.0$ & $619.8$ & $587.5\,(32.3)$ & $374.4$ \\
 & SA $\downarrow$ & $2.31$ & $5.70$ & $3.91$ & $4.25$ & $4.00$ & $4.03\,(0.05)$ & $3.10$ \\
 & Corr & $0.30$ & $0.75$ & $0.77$ & $0.82$ & $0.77$ & $0.69$ & $0.34$ \\
 & $p$ & $<0.001$ & $<0.001$ & $<0.001$ & $<0.001$ & $<0.001$ & $<0.001$ & $<0.001$ \\
\midrule
\multirow{4}{*}{QED} & MW $\downarrow$ & $274.7$ & $317.2$ & $290.4$ & $255.0$ & $265.2$ & $300.8\,(8.8)$ & $270.8$ \\
 & SA $\downarrow$ & $2.04$ & $3.18$ & $3.85$ & $3.09$ & $3.02$ & $3.02\,(0.18)$ & $2.34$ \\
 & Corr & $0.32$ & $0.40$ & $0.52$ & $0.39$ & $0.42$ & $0.16$ & $0.06$ \\
 & $p$ & $<0.001$ & $<0.001$ & $<0.001$ & $<0.001$ & $<0.001$ & $<0.001$ & $0.008$ \\
\midrule
\multirow{4}{*}{sEH} & MW $\downarrow$ & $268.9$ & $426.6$ & $632.6$ & $524.4$ & $574.2$ & $440.9\,(11.1)$ & $278.7$ \\
 & SA $\downarrow$ & $2.17$ & $3.36$ & $3.66$ & $3.10$ & $3.38$ & $3.13\,(0.30)$ & $2.38$ \\
 & Corr & $0.47$ & $0.52$ & $0.77$ & $0.63$ & $0.75$ & $0.34$ & $-0.04$ \\
 & $p$ & $<0.001$ & $<0.001$ & $<0.001$ & $<0.001$ & $<0.001$ & $<0.001$ & $0.044$ \\
\bottomrule
\end{tabular}
\end{center}
\makeatother

While PGFS++ demonstrates strong capability in optimizing target properties, it does not consistently outperform all baselines in molecular weight. In the HIV-Int optimization task, PGFS++ generated
molecules with a mean molecular weight of 415~Da (Fig.~\ref{fig:supp-mw}), comparable to PGFS and PGFS MB, lower than SynFlowNet and PGFS TAC, and higher than SynFormer and REINVENT4. A similar trend holds for ChEMBL out-of-distribution test molecules, where PGFS++ yields an average molecular weight of \(587.5\)~Da after HIV-Int optimization, slightly exceeds the 500~Da molecular-weight guideline in Lipinski's rule~\cite{lipinski1997experimental}. Nevertheless, the mean molecular weight remains well within the range reported for approved oral drugs and clinical candidates, some of which are orally bioavailable at molecular weights of 700--1,000~Da~\cite{doak2014oral,shultz2019two}. Overall, although PGFS++ does not consistently outperform all baselines in molecular weights, its generated molecules are predicted to have high target-property scores and reasonable molecular weights.

These differences in output molecular weights may partly explain why PGFS++ achieved lower $1-\mathrm{SA}$ scores than some baselines. Among PGFS++ outputs, heavier molecules tended to have higher SA scores and thus lower $1-\mathrm{SA}$ scores (Table~\ref{tab:supp-mw-sa}). This association was strongest on the out-of-distribution ChEMBL task (Spearman's $\rho=0.69$). On ChEMBL, both SynFormer and REINVENT4 generated lighter molecules with lower mean SA scores than PGFS++. However, PGFS generated slightly heavier molecules than PGFS++ on average (598.7 versus 587.5~Da) and achieved a lower mean SA score (3.91 versus 4.03). Molecular weight therefore offers a partial explanation for the differences in synthetic accessibility across methods.

\clearpage
\subsection{Molecular examples of input--output changes}
\label{sec:supp-route-examples}
\begin{figure}[p]
\centering
\makebox[\textwidth][c]{\includegraphics[width=1.32\textwidth]{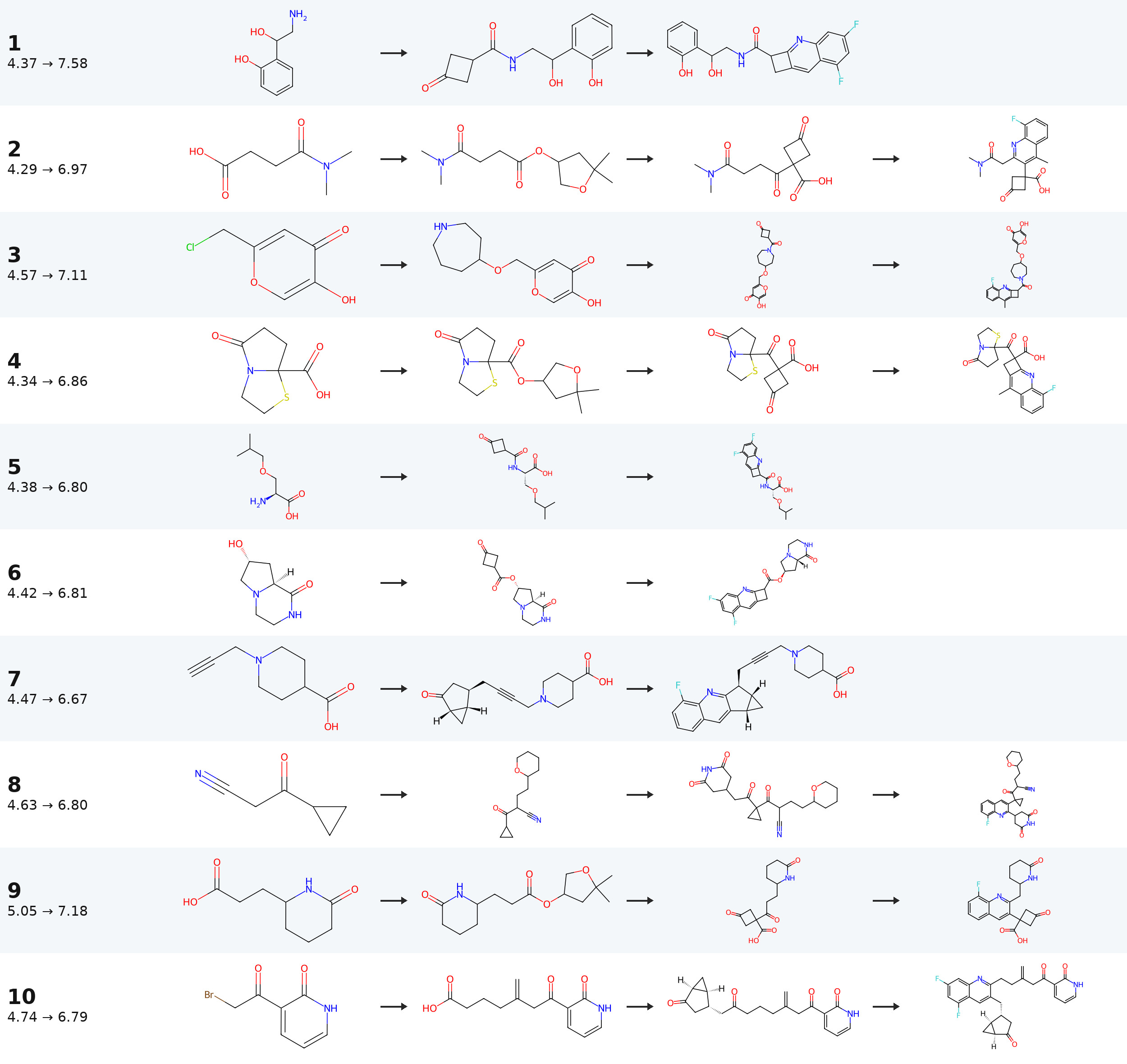}}
\caption{Ten example reaction routes for HIV-Int optimisation, showing the starting molecules, intermediate products where present, and final products. Values beneath each route number indicate the predicted $\mathrm{pIC}_{50}$ of the starting molecule and final product, respectively (input score $\rightarrow$ output score).}
\label{fig:supp-routes-hiv}
\end{figure}

\begin{figure}[p]
\centering
\makebox[\textwidth][c]{\includegraphics[width=1.32\textwidth]{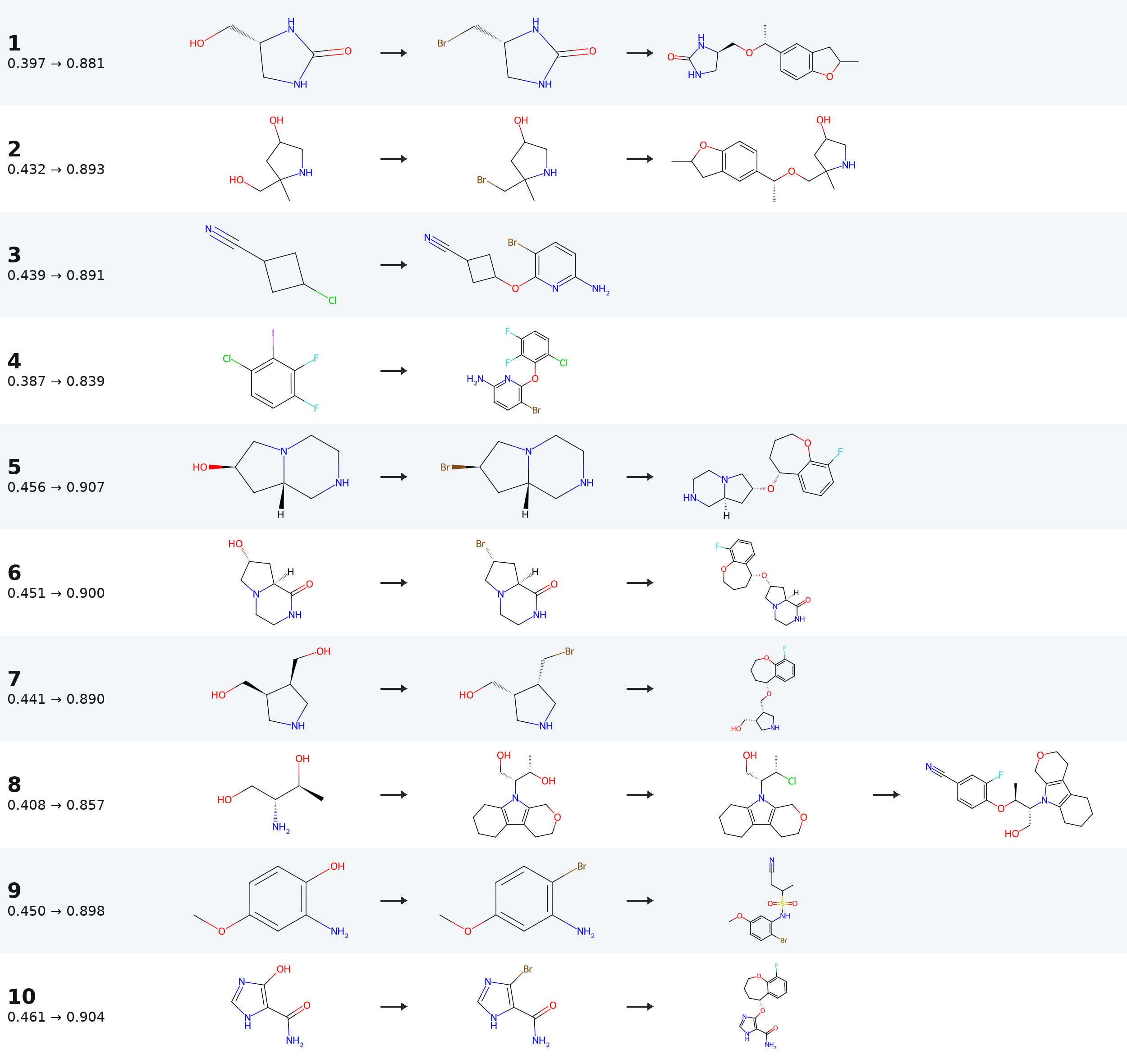}}
\caption{Ten example reaction routes for QED optimisation, showing the starting molecules, intermediate products where present, and final products. Values beneath each route number indicate the QED of the starting molecule and final product, respectively (input score $\rightarrow$ output score).}
\label{fig:supp-routes-qed}
\end{figure}

\begin{figure}[p]
\centering
\makebox[\textwidth][c]{\includegraphics[width=1.32\textwidth]{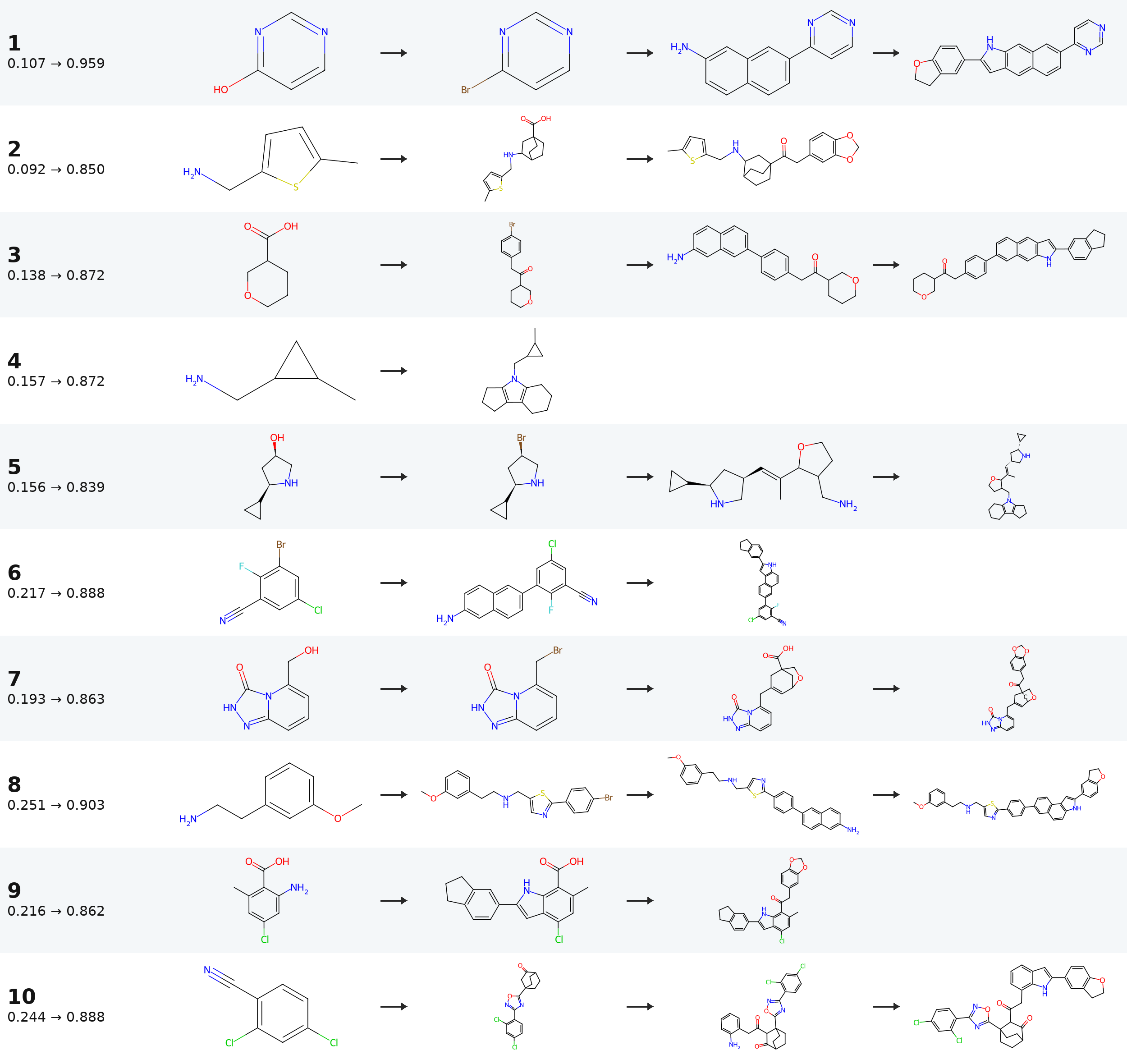}}
\caption{Ten example reaction routes for sEH optimisation, showing the starting molecules, intermediate products where present, and final products. Values beneath each route number indicate the sEH score of the starting molecule and final product, respectively (input score $\rightarrow$ output score).}
\label{fig:supp-routes-seh}
\end{figure}

Figures~\ref{fig:supp-routes-hiv}--\ref{fig:supp-routes-seh} show sample PGFS++ routes for HIV-Int, QED and sEH. Each row contains the input, any intermediate products and the output, and an arrow denotes one reaction. The number beside a row is the objective score of the input followed by the score of the output.

\clearpage
\subsection{Magnet-output routes}
\label{sec:supp-magnet-routes}
\begin{figure}[p]
\centering
\makebox[\textwidth][c]{\includegraphics[width=1.32\textwidth]{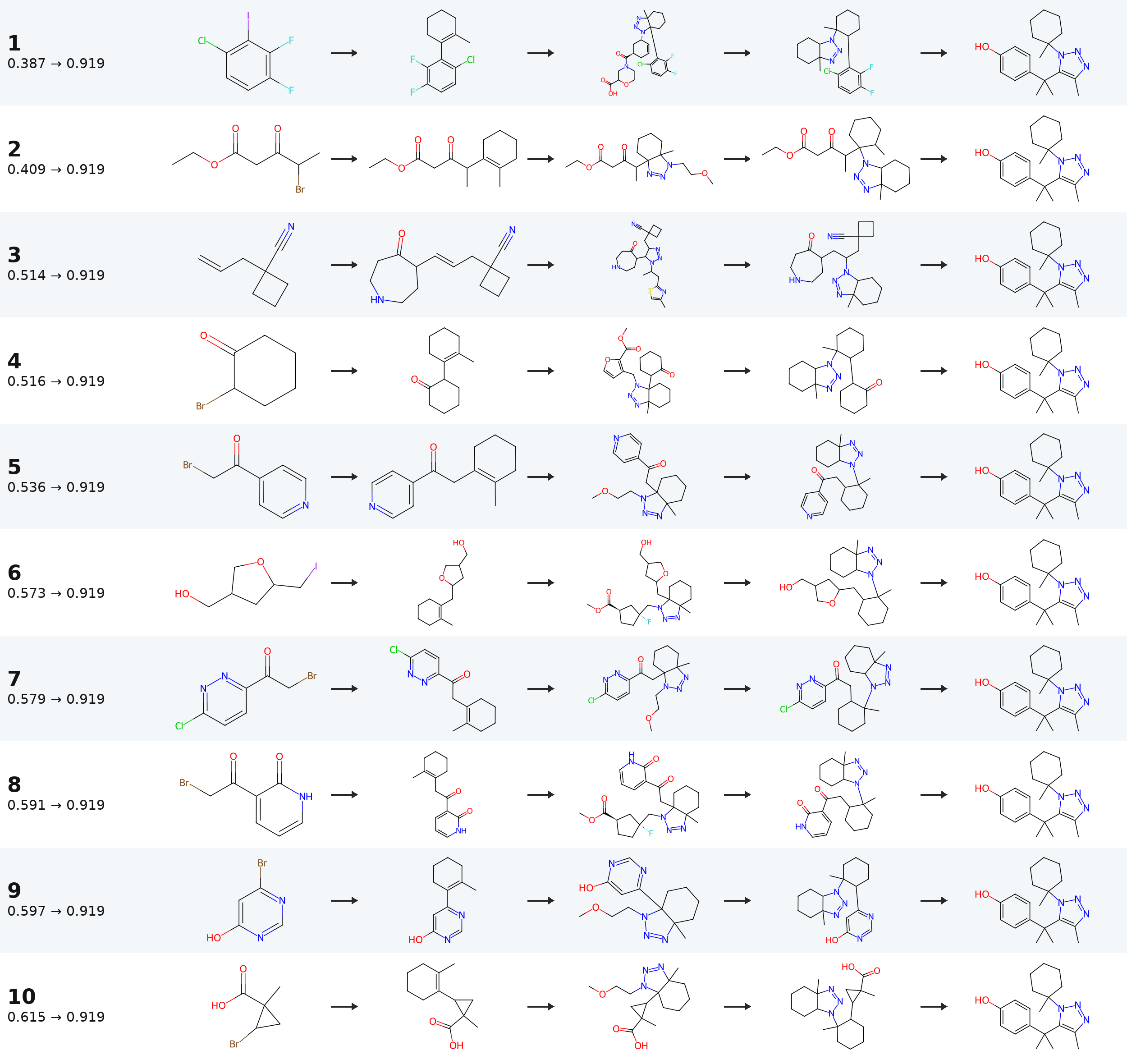}}
\caption{Representative magnet molecules and the reaction routes generated by PGFS+ for QED optimisation task.}
\label{fig:supp-magnet-pgfs}
\end{figure}

\begin{figure}[p]
\centering
\makebox[\textwidth][c]{\includegraphics[width=1.32\textwidth]{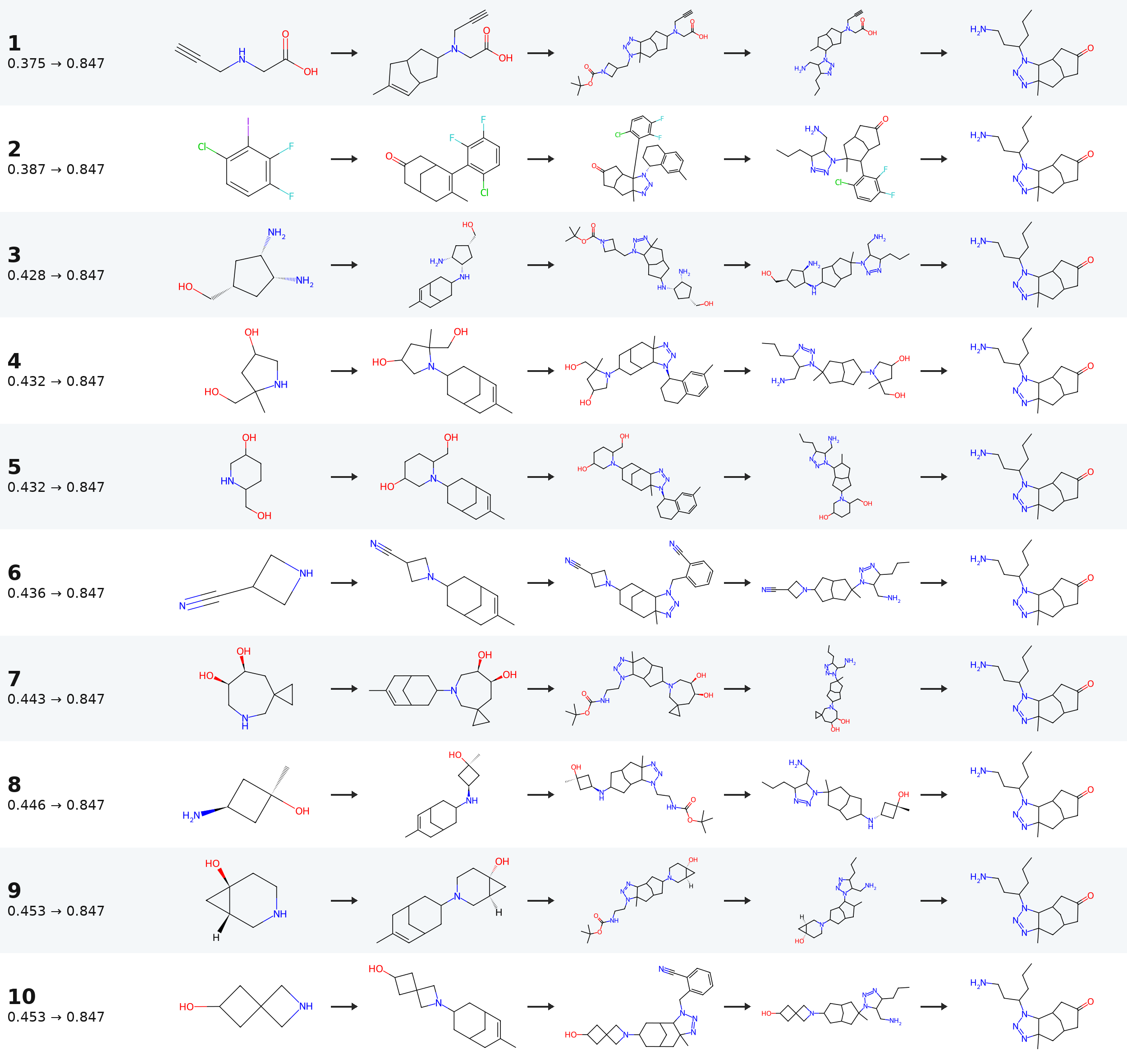}}
\caption{Representative magnet molecules and the reaction routes generated by avoid-list penalty for QED optimisation task.}
\label{fig:supp-magnet-avoid}
\end{figure}

\begin{figure}[p]
\centering
\makebox[\textwidth][c]{\includegraphics[width=1.32\textwidth]{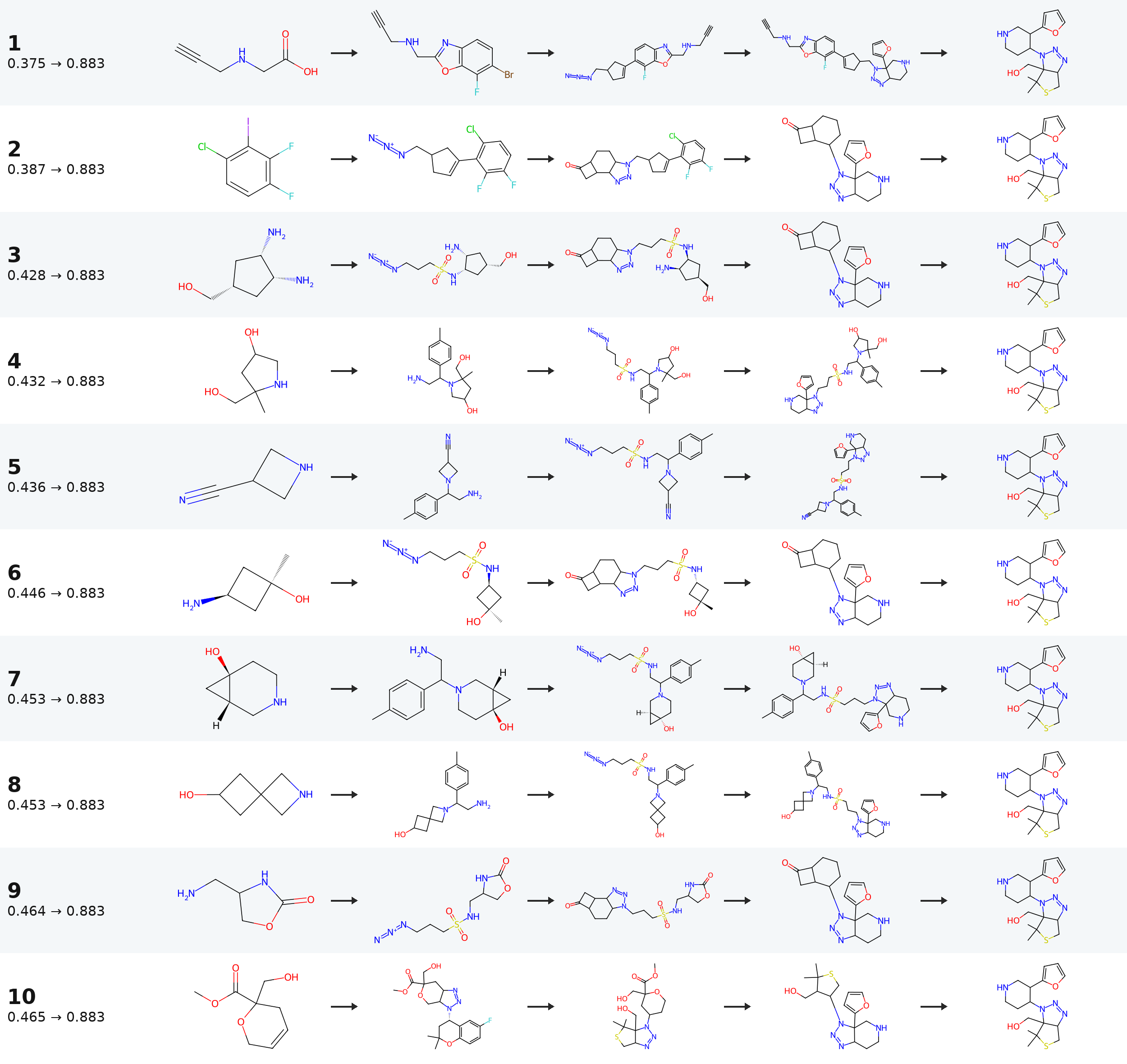}}
\caption{Representative magnet molecules and the reaction routes generated by soft penalty for QED optimisation task.}
\label{fig:supp-magnet-soft}
\end{figure}

Figures~\ref{fig:supp-magnet-pgfs}--\ref{fig:supp-magnet-soft} show ten routes, each of at most four reactions, that end on the magnet molecule. The routes are taken from the $2{,}000$-molecule test set.

\clearpage
\bibliography{references}